\documentclass[letterpaper, 10 pt, journal, twoside]{IEEEtran}

\usepackage[T1]{fontenc}

\usepackage{amsfonts}       % blackboard math symbols
\usepackage{nicefrac}       % compact symbols for 1/2, etc.
\usepackage{subfig}			% per usare le subfigure
\usepackage{booktabs}	% toprule midrule bottomrule
\usepackage[linesnumbered,ruled,vlined]{algorithm2e}
\usepackage{xcolor}
\usepackage[normalem]{ulem} % either use this (simple) or
\usepackage{mathtools}
\usepackage[percent]{overpic} 
\usepackage{flushend}
\usepackage{comment}
\usepackage[noadjust]{cite}

\definecolor{black}{rgb}{0, 0, 0}
\definecolor{red}{rgb}{0.9, 0, 0}
\definecolor{green}{rgb}{0, 0.6, 0}
\definecolor{blue}{rgb}{0, 0, 0.9}
\definecolor{grey}{rgb}{0.52, 0.52, 0.51}

\newcommand{\RED}[1]{\textcolor{red}{#1}}

\newcommand{\TODO}[1]{\RED{\textbf{TODO: } #1}}

\SetKwInOut{KwParameter}{Parameters}
\SetKw{KwContinue}{continue}
\SetKw{KwBreak}{break}
\SetKw{KwInterval}{in}
\SetKw{KwOr}{or}
\SetKw{KwAnd}{and}
\SetKwFunction{unif}{unif}

\newcommand{\X}[0]{X}
\newcommand{\x}[0]{x}
\newcommand{\U}[0]{U}
\renewcommand{\u}[0]{u}

\newcommand{\fillbox}[3]% #1=width, #2=height, #3=filename
{\bgroup
  \dimen1=#1\relax% store width into register
  \dimen2=#2\relax% store height into register
  \sbox0{\includegraphics[width=#1]{#3}}%
  \ifdim\ht0>\dimen2
    \dimen0=\dimexpr \ht0-\dimen2\relax
    \adjustbox{clip=true,trim=0pt 0.5\dimen0 0pt 0.5\dimen0}{\usebox0}%
  \else
    \sbox0{\includegraphics[height=#2]{#3}}%
    \ifdim\wd0>\dimen1
      \dimen0=\dimexpr \wd0-\dimen1\relax
      \adjustbox{clip=true,trim=0.5\dimen0 0pt 0.5\dimen0 0pt}{\usebox0}%
    \else
      \usebox0
    \fi
  \fi
\egroup}

\newcommand{\parent}{p}
\newcommand{\child}{c}
\newcommand{\target}{\mathrm{trg}}

\DeclareMathOperator*{\argmax}{argmax}

\begin{document}

\setlength{\textfloatsep}{10pt}% Remove \textfloatsep

\title{Motion Planning as Online Learning: \\ A Multi-Armed Bandit Approach to Kinodynamic Sampling-Based Planning}

\author{Marco Faroni and Dmitry Berenson %,~\IEEEmembership{Senior~Member,~IEEE}
%\thanks{Manuscript received: March, 24, 2023; Revised May, 22, 2023; Accepted July, 29, 2023.  This paper was recommended for publication by Editor Hanna Kurniawati upon evaluation of the Associate Editor and Reviewers' comments. }
%\thanks{This work was supported in part by Toyota Research Institute, the Office of Naval Research Grant N00014-21-1-2118, and %NSF grants IIS-1750489, IIS-2113401, and IIS-2220876. }
\thanks{The authors are with the Robotics Department, University of Michigan, Ann Arbor, MI 48109, United States. {\tt\footnotesize \{mfaroni; dmitryb\}@umich.edu}}
\thanks{Digital Object Identifier (DOI): see top of this page.}}

\maketitle

\markboth{IEEE Robotics and Automation Letters. Preprint Version. Accepted July, 2023}
{Faroni \MakeLowercase{\textit{et al.}}: Motion Planning as Online Learning: A Multi-Armed Bandit Approach} 

\begin{abstract}
Kinodynamic motion planners allow robots to perform complex manipulation tasks under dynamics constraints or with black-box models.
However, they struggle to find high-quality solutions, especially when a steering function is unavailable.
This paper presents a novel approach that adaptively biases the sampling distribution to improve the planner's performance.
The key contribution is to formulate the sampling bias problem as a non-stationary multi-armed bandit problem, where the arms of the bandit correspond to sets of possible transitions.
High-reward regions are identified by clustering transitions from sequential runs of kinodynamic RRT and a bandit algorithm decides what region to sample at each timestep.
The paper demonstrates the approach on several simulated examples as well as a 7-degree-of-freedom manipulation task with dynamics uncertainty, suggesting that the approach finds better solutions faster and leads to a higher success rate in execution.
\end{abstract}


\IEEEpeerreviewmaketitle

\section{Introduction}\label{sec:intro}

\begin{figure}[tpb]
	\centering
	\includegraphics[trim = 38cm 9.7cm 20cm 4.3cm, clip, angle=0, width=0.69\columnwidth]{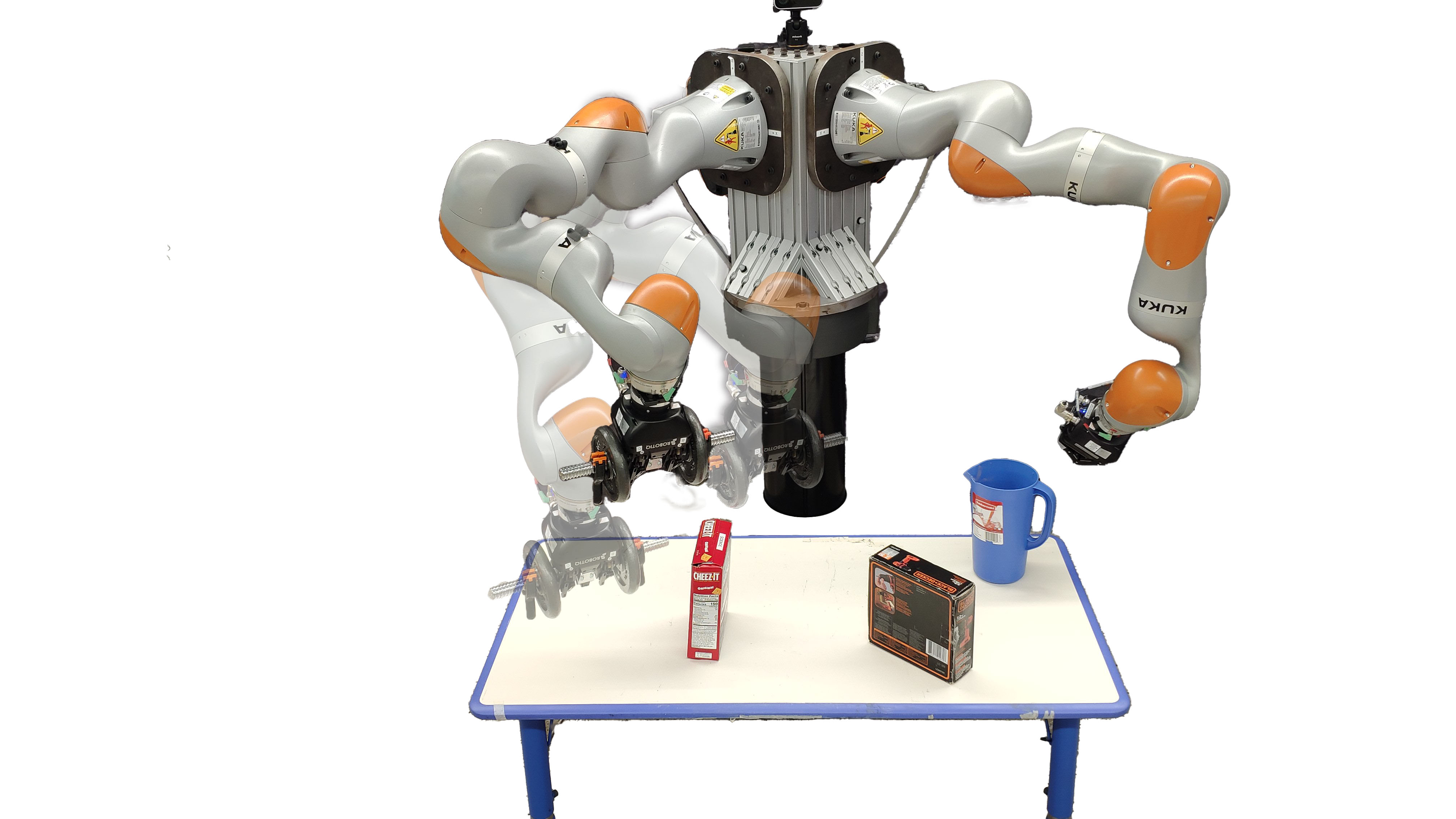}
	\caption{A 7-degree-of-freedom manipulator carrying a weight in a cluttered environment with uncertain tracking control.}
	\label{fig: victor-dumbbell}
\end{figure}

\IEEEPARstart{P}{hysics} simulators and deep-learning models allow robots to reason about complex manipulation tasks such as manipulation of deformable objects \cite{Mitrano:science-robotics,corl-planning-with-spatial-temporal-abstraction,Lippi:visual-action-planning}, liquid handling \cite{Mitrano:focus-adaptation,fluid-manipulation}, and contact-rich manipulation \cite{Kromere:contact-rich-manipulation,motion-planning-sliding}.
Kinodynamic motion planning can find a sequence of controls that brings such systems to a desired state.
For example, consider the tabletop scenario in Fig. \ref{fig: victor-dumbbell}: A compliant manipulator moves a heavy object across the table; because of the payload and the compliant control, the trajectory execution will deviate from the planned path, possibly causing unexpected collisions.
Suppose we have a function that maps the robot state to an estimate of the end-effector Cartesian error; we can avoid unexpected collisions in execution by finding a trajectory that minimizes such a function.
%Similar considerations apply to different domains, such as planning for deformable objects \cite{Mitrano:science-robotics}, human-aware motion planning \cite{mainprice2011planning}, or navigation on uneven terrains \cite{rrt-uncertain-terrain}.

Sampling-based planners, such as rapidly exploring random trees (RRT) \cite{lavalle:rrt-ijrr} 
%and probabilistic roadmaps (PRM) \cite{Kavraki:PRM}
are widely used in robotics because of their effectiveness in high-dimensional problems.
%They search for a valid path by building a tree that extends toward sampled states.
Despite the fact that asymptotically optimal algorithms \cite{karaman:RRT*,kinodynamic-rrt-star} ensure convergence to the optimal solution for an infinite number of iterations, their convergence rate is often slow for practical applications.
This issue holds especially if a steering function (\emph{i.e.}, a function that connects two given states) is not available or computationally expensive, which is often the case for learned or simulated dynamics models \cite{bekris:sst}.

%Many planners ensure  asymptotic convergence to the optimal trajectory if a steering function (\emph{i.e.}, a function that connects two given states) is available \cite{karaman:RRT*,kinodynamic-rrt-star}.
%However, computing a steering function  %requires solving a two-point bounded-value problem, which is often intractable or computationally expensive.
%In the case of black-box functions such as learned or simulated dynamics models, the steering function might not be available at all.
%\rev{To deal with that, existing planners rely only on forward propagation to extend the tree with sample random controls and duration \cite{bekris:sst} or run kinodynamic RRT multiple times \cite{Hauser:AO-RRT}.}
%In these cases, the convergence rate of existing algorithms is often too slow for practical applications. 

Planning performance can be improved by biasing the sampling distribution, \emph{e.g.}, to find a solution faster \cite{heuristic-rrt} or  to reduce the cost of the solution \cite{Gammel:InformedRRT}.
RRT-like planners with biased sampling extend the tree by sampling a target state from a non-uniform distribution.
The biased distribution can be learned offline \cite{Pavone:learning-sampling,Kingston:learning-sampling,Burdick:learning-sampling-distribution} or adapted online based on previous iterations \cite{gammell:rabit,Gammell2015-bit}.
We approach the problem of biased sampling from an online-learning perspective.
That is, we consider biased sampling as a sequential decision-making process where each transition added to the tree is associated with a reward (dependent on the cost function).
Then, we decide what transition to sample at the next iteration based on the rewards estimated from previous timesteps.
In particular, this paper proposes an online learning approach to biasing samples in a kinodynamic RRT.

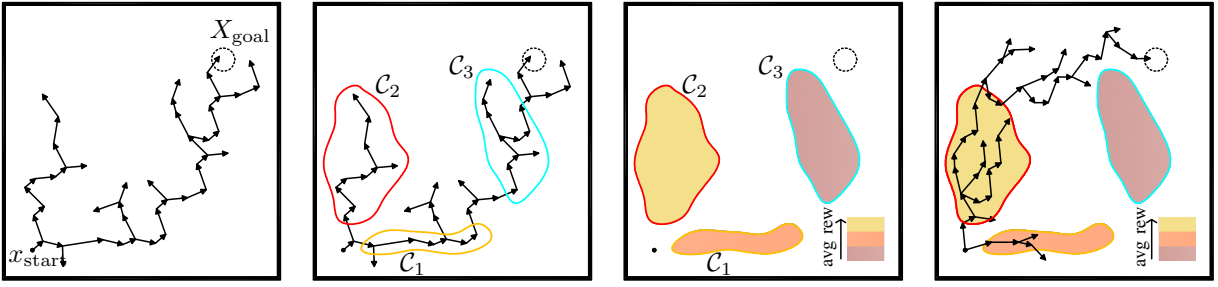
\begin{figure*}[tbp]
	\centering
    % grid,tics=5
    \begin{overpic}[width=0.9\textwidth]{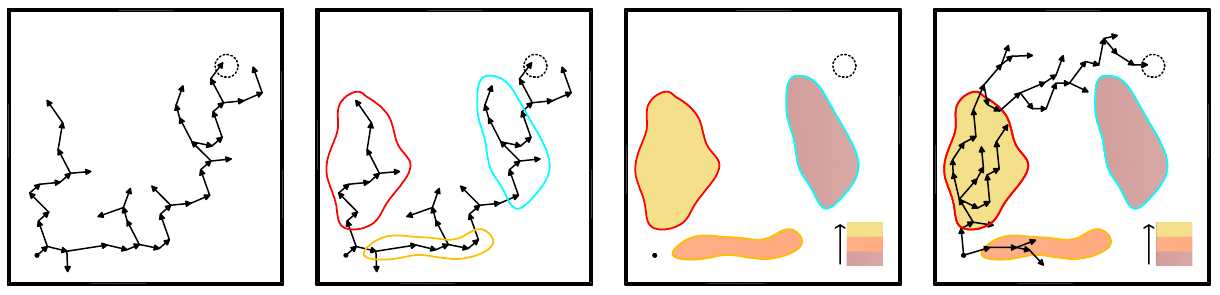} 
    \put (1.1,2.2) {$\x_{\mathrm{start}}$}
    \put (17.5,20.5) {$X_{\mathrm{goal}}$}
    \put (33,1.5) {$\mathcal{C}_1$}
    \put (31,16) {$\mathcal{C}_2$}
    \put (37,17.5) {$\mathcal{C}_3$}
    \put (58,1.5) {$\mathcal{C}_1$}
    \put (56,16) {$\mathcal{C}_2$}
    \put (62.2,17.5) {$\mathcal{C}_3$}
    \put (67.5,1.8) {\rotatebox{90}{\scriptsize avg rew}}
    \put (93,1.8) {\rotatebox{90}{\scriptsize avg rew}}
    \end{overpic}
	\caption{Sketch of the proposed method. From left to right: (i) a randomized planner searches for a solution without bias; (ii) each time it finds a new solution, transitions are clustered based on their reward and position; (iii) each cluster is associated with an estimated reward; (iv) a non-stationary bandit algorithm biases sampling based on the regions' expected reward.}
	\label{fig: example}
\end{figure*}

We use Multi-Armed Bandit (MAB) algorithms to shape the sampling probability distribution iteratively.
The proposed method is illustrated in Fig. \ref{fig: example}.
Our approach builds on the asymptotically optimal framework AO-RRT \cite{Hauser:AO-RRT}, which runs kinodynamic RRT multiple times. Every time a new solution is found, transitions are clustered based on their reward and spatial position, and a non-stationary bandit algorithm biases samples based on the expected reward of each region.

The contributions of this paper are:
\begin{itemize}
    \item An online learning approach to biasing samples in a motion planner that formulates the bias problem as a non-stationary Multi-Armed Bandit problem and trades off the exploration and exploitation of high-reward regions based on the reward observed at previous timesteps. 
    \item A kinodynamic planner that does not rely on a steering function and uses the proposed MAB approach to find better solutions faster. 
    \item A demonstration of the proposed method on a 7-degree-of-freedom manipulation problem (Fig. \ref{fig: victor-dumbbell}), showing that the proposed approach improves the solution cost  and, in the scenario at hand, leads to a higher success rate in execution.
    An empirical regret analysis of different sampling strategies also suggests that a better solution cost coincides with a lower cumulative regret.
\end{itemize}

%The paper is organized as follows:
%Sec. \ref{sec:related-works} discusses the related works, Sec. \ref{sec:preliminaries} formulates the adaptive sampling problem, Sec. \ref{sec:method} presents the proposed approach, Sec.s \ref{sec:numerical} and \ref{sec:experiments} show numerical and experimental results, and Sec. \ref{sec:conclusions} draws the conclusions.
%The results of the experiments are shown in the video attached to the manuscript.

\section{Related Work} \label{sec:related-works}

In the last decade, sampling-based motion planning has seen a focus shift from finding feasible solutions to finding high-quality ones, especially after \cite{karaman:RRT*} provided conditions for asymptotic optimality for planners such as RRT$^*$ and PRM$^*$.
These conditions include the availability of a steering function, making these planners unsuitable for black-box dynamics models.
%However, kinodynamic motion planning is a harder problem than its kinematic counterpart as it involves higher dimensional spaces and dynamics constraints \cite{gammell:review}.
%Sampling-based planners are well suited to this problems as they generalize well to different systems and scale to high-dimensional problems.
To overcome this issue, variants of RRT$^*$ have been proposed by approximating the steering function \cite{kinodynamic-rrt-star,Frazzoli:approximated-steering-function,steering-approximation}, but these approaches are only suitable for a limited class of systems.
Other works researched how to guarantee asymptotic optimality without a steering function \cite{bekris:sst,Hauser:AO-RRT}.
In particular, \cite{Hauser:AO-RRT} proposed an asymptotically optimal meta-planning algorithm based on multiple runs of RRT in an augmented state-cost space, whereas \cite{bekris:sst} combined biased node expansion with pruning to refine an initial solution.
Attempts to improve the convergence rate of these methods include using a heuristic to bias node expansion \cite{Bekris:DIRT,Ruml:RIOT}, pruning and re-usage of previous edges \cite{Bekris:proof-kinodynamic-rrt}, and building a PRM-like roadmap of edges offline \cite{Kavraki:bundle-of-edges}.

In this work, we focus on improving the solution quality of kinodynamic planners via adaptive sampling. 
Most sampling-based planners use a uniform sampling distribution; however, biasing the sampling has been a common strategy to improve the solution cost \cite{gammell:review,amato:obprm,amato:evaluating-guiding-spaces,Pavone:learning-sampling,Burdick:learning-sampling-distribution,Kingston:learning-sampling,Laumond:pca-rrt}. 
The sampling bias is often tailored to the specific problem manually  \cite{suarez:task-vector-rrt} or using machine learning techniques \cite{Pavone:learning-sampling,Kingston:learning-sampling,Burdick:learning-sampling-distribution}.
Other works leverage the knowledge gathered at the previous iterations to bias the sampling at the current one.
For example, they use local bias to overcome narrow passages \cite{ramos:multi-tree-mab-RAL}, switch between global and local sampling to find a solution faster \cite{RRT-bayesian-sampling}, or quickly refine the current solution \cite{gammell:rabit,Gammell-ABIT}.
The works above are designed for holonomic motion planners.
Informed sampling is another biased sampling technique that uses cost heuristics to discard regions with a null probability of improving previous solutions \cite{Gammel:InformedRRT}.
How to derive or approximate such heuristics for non-trivial cost functions is an open research question \cite{Gammell-AIT,Gammell-ABIT}.

Our approach leverages MAB algorithms to choose the sampling bias online.
MAB is an online learning technique used for repeated decision-making under uncertainty.
An MAB problem is defined by a set of actions (\emph{arms}) associated with a belief of their reward function.
At each iteration, an agent chooses an arm and updates the reward estimates according to its realized reward.
MAB algorithms are typically characterized by their regret, \emph{i.e.}, how much worse they perform compared to a strategy that picks the best arm at each iteration.
Different approaches (and regret bounds) have been derived based on different assumptions on the reward distribution.
Common algorithms are UCB-1 \cite{UCB1} and Thompson Sampling \cite{thompson} for constant reward distributions and their variants for non-stationary rewards \cite{kfmanb}.
A comprehensive overview of the topic can be found in \cite{slivkins:mab-introduction,lattimore:bandit-book}.

Recent works applied the MAB framework to motion planning, aiming to automatically balance the trade-off between exploration and exploitation \cite{Likhachev:mab-ensemble-heuristics,Srinivasa:pareto-optimal,Srinivasa:mab-rearrangement-planning}.
%For example, \cite{Likhachev:mab-ensemble-heuristics} used a bandit algorithm to dynamically schedule different heuristics in graph-based search;
%\cite{Srinivasa:pareto-optimal} used a bandit algorithm for multi-objective trajectory optimization; and
%\cite{Srinivasa:mab-rearrangement-planning} optimized the selection of trajectories for rearrangement planning.
To the best of our knowledge, MAB was applied to sampling-based planning only to overcome narrow passages in bi-directional search \cite{LM-RRT,ramos:multi-tree-mab-RAL}.
They consider trees as bandits' arms and decide which tree to expand depending on the estimated probability of a successful expansion.
%Both consider search with multiple trees and associate a tree with a bandit's arms so that the decision of which tree to extend at each iteration depends on the estimated probability of successful extension.
However, this technique is specific to narrow passages and works only with multiple trees and the availability of a steering function. 

\section{Problem Statement}\label{sec:preliminaries}

Consider a dynamics model $\dot{x} = f(\x, \u)$,  $\x \in \X$,  $\u \in \U$, where $\X$ and $\U$ are the state and the control spaces and $\X_{\mathrm{free}} \subseteq \X$ is the set of valid states.
Solving a kinodynamic motion planning problem means finding a 
%sequence of controls  and durations $\{(u_i,d_i)\}$ 
control function $\gamma: [0,T] \in \U$
that induces a trajectory $\sigma: [0,T] \rightarrow \X$ such that $\sigma(0) = \x_{\mathrm{start}}$ and $\sigma(T) \in X_{\mathrm{goal}}$, and $\sigma(t) \in \X_{\mathrm{free}} \, \forall t \in [0,T]$.
In optimal motion planning, we also aim to minimize a Lipschitz continuous cost function $c(\sigma)$.
In this work, we restrict $\gamma$ to be a staircase function defined by a sequence of controls and control durations, $\{ (u_i,d_i) \}$, so that $\sum_{i} d_i = T$ and 
$
\gamma(t) = u_j \text{ with } j\in \mathbb Z^+ | \textstyle\sum_{i=0}^j d_i \leq t_j < \textstyle\sum_{i=0}^{j+1} d_i
$.

Sampling-based planners such as kinodynamic RRT solve this problem by randomly sampling a target state, $\x_{\target}$, retrieving the closest node on the tree, $\x_{\parent}$, and expanding this node by forward propagation.
We denote by $\tau = (\x_{\parent}, \u, d, \x_{\child},\x_{\target})$ the transition from $\x_{\parent}$ to $\x_{\child}$ induced by $\u$ and $d$; note that $\tau$ also stores the target $\x_{\target}$ from which $\tau$ originated.
%The resulting state, $\x_{\child}$, defines a new transition, $\tau = (\x_{\parent},\x_{\child},\x_{\target})$, that is attached to the tree.
%\rev{Depending on the problem specifications and the implementation of the state and control sampling, they may guarantee probabilistic completeness and asymptotic optimality \cite{stilman:rrt-not-complete,bekris:sst}.}

The search strategy above is a sequential decision-making process, where the planner has to choose which node to expand next and in what direction.
We consider this process from an online-learning perspective, where each new transition is associated with a reward $r = \rho(\tau)$, where $0\leq\rho\leq 1$\footnote{Note that $\rho(\tau)$ should be inversely proportional to the cost of $\tau$; however, its definition may be problem-specific. We propose examples of the reward function in Sec. \ref{sec:numerical} and \ref{sec:experiments}.}.
The planner observes the reward of a transition after each iteration and chooses the next transition to maximize the total reward over $K$, possibly infinite, timesteps.
MAB is a framework to address this kind of problems.
In the MAB settings, an agent can choose among $M$ actions (the \emph{arms}) for $K$, possibly infinite, rounds.
The goal is to maximize the cumulative reward, assuming each action yields a reward from an unknown distribution and the agent can only observe the reward of the selected action.

We frame the problem of choosing the next transition in a kinodynamic RRT as an MAB problem where the arms are sets of transitions. Selecting an arm then corresponds to sampling a transition $\tau$ from a certain set and using it to extend the tree.
Our goal is to use MAB to improve the path cost over iterative runs of kinodynamic RRT by trading off the exploitation of high-reward regions (according to the reward obtained at previous runs) and the exploration of transitions with a highly uncertain reward (\emph{i.e.}, less-explored regions).

\section{Method}\label{sec:method}

\begin{algorithm}[tpb]
\DontPrintSemicolon
\SetKwFunction{sample}{sample}
\SetKwFunction{sampleAndPropagate}{sampleAndPropagate}
\SetKwFunction{add}{add}
\SetKwFunction{updateBanditArms}{updateBanditArms}
\SetKwFunction{initializeBandits}{initializeBandits}
\SetKwFunction{retracePath}{retracePath}
\SetKwFunction{clustering}{clustering}
\SetKwFunction{sampleTo}{sampleTo}
\KwIn{$\x_{\mathrm{start}}$, $X_{\mathrm{goal}}$, $c(\cdot)$, $\rho(\cdot)$, $K>0$}
\KwOut{$\sigma_{\mathrm{best}}$, $c_{\mathrm{best}}$}
$\sigma_{\mathrm{best}} \leftarrow \emptyset$, $c_{\mathrm{best}} \leftarrow +\infty$, $\mathcal{T} \leftarrow \emptyset$\;
$\mathcal{G} \leftarrow \emptyset$, $\mathcal{G}.\add(\x_{\mathrm{start}})$\;
$\RED{\initializeBandits(0,0)}$\; \label{alg:MAB-RRT:initMAB}
\For{$k$ \KwInterval $1,\dots,K$}
{
     $\RED{\tau \leftarrow {\sampleAndPropagate(\mathcal{C}, \mathcal{G})}}$\; \label{alg:MAB-RRT:sample}
     $\RED{\updateBanditArms(\rho(\tau))}$\; \label{alg:MAB-RRT:updateMAB}
     $\mathcal{G}.\add(\tau)$\; \label{alg:MAB-RRT:add-to-nn}
     \If{$x_\mathrm{next} \in X_{\mathrm{goal}}$}{ \label{alg:MAB-RRT:goal-check}
        $\sigma \leftarrow \retracePath(\tau)$\; \label{alg:MAB-RRT:find-path}
        \If{$c(\sigma) < c_{\mathrm{best}}$}{
            $\sigma_{\mathrm{best}} \leftarrow \sigma$, \, $c_{\mathrm{best}} \leftarrow c(\sigma)$\;
        }   
        ${\mathcal{T} \leftarrow \mathcal{T} \cup \mathcal{G}}$\;   \label{alg:MAB-RRT:clustering-1}
        $\RED{\mathcal{C}, \mathcal{R} \leftarrow \clustering(\mathcal{T})}$\;  \label{alg:MAB-RRT:clustering-2}
        $\RED{\initializeBandits(\mathcal{R}_1,\dots,\mathcal{R}_M)}$\;  \label{alg:MAB-RRT:reinitMAB}
        $\mathcal{G} \leftarrow \emptyset,\, \mathcal{G}.\add(\x_{\mathrm{start}})$\; \label{alg:MAB-RRT:resetRRT}
     }      
 }
 \Return $\sigma_{\mathrm{best}}$, $c_{\mathrm{best}}$\;
 \caption{\FuncSty{MAB-RRT}}
 \label{alg:MAB-RRT}
\end{algorithm}

Our approach can be summarized as follows: (a) we iteratively re-plan with kinodynamic RRT, using MAB to select regions for sampling transitions; (b) every time we find a new solution trajectory, we identify high-reward regions by clustering previous transitions; (c) our MAB method estimates the non-stationary reward distribution as we plan during a run of kinodynamic RRT.
This section describes the planning framework and methods for clustering and biased sampling.

\subsection{Planning framework}

We propose a kinodynamic planner based on the AO-RRT meta-planning paradigm \cite{Hauser:AO-RRT}, which runs instances of RRT sequentially and keeps the best solution so far. 
Alg. \ref{alg:MAB-RRT} summarizes the proposed algorithm (the differences with respect to AO-RRT are in red).
At each iteration, \texttt{sampleAndPropagate} (line \ref{alg:MAB-RRT:sample}) samples a transition.
The new transition is added to the tree (line \ref{alg:MAB-RRT:add-to-nn}).
If the goal condition is satisfied, the solution is retrieved  by \texttt{retracePath} (lines \ref{alg:MAB-RRT:goal-check}--\ref{alg:MAB-RRT:find-path}) and RRT is reset (line \ref{alg:MAB-RRT:resetRRT}).
After $K$ iterations, Alg. \ref{alg:MAB-RRT} returns the best solution so far, $\sigma_{\mathrm{best}}$, and its cost, $c_{\mathrm{best}}$.
We embed our adaptive sampling strategy in AO-RRT in two steps: (a) clustering and (b) bandit-based sampling.

\subsubsection{Clustering}

At the end of each RRT run, we add all the transitions to the set of all previous transitions, $\mathcal{T}$,  and cluster them into a set of clusters $\mathcal{C}$ (lines \ref{alg:MAB-RRT:clustering-1} and \ref{alg:MAB-RRT:clustering-2}) by using HDBSCAN \cite{HDBSCAN}.
Then, we associate each cluster $\mathcal{C}_i$ with a bandit's arm and use each cluster's average reward, $\mathcal{R}_i$, to initialize the bandit's expected rewards (line \ref{alg:MAB-RRT:reinitMAB}).

\subsubsection{Bandit-based sampling}

An MAB algorithm in \texttt{sampleAndPropagate} decides whether to sample the next transition $\tau$ from a cluster, the uniform distribution, or the goal set.
Then, we extend the tree from $\tau.\x_{\parent}$ by forward propagation.
After the extension, the MAB updates the arms' rewards according to the reward realized by the new transition (line \ref{alg:MAB-RRT:updateMAB}).
Note that the reward function is non-stationary with respect to the tree, as detailed in Sec.\ref{sec:adaptive-sampling} and \ref{sec:non-stationary}, because the transition reward depends on the current state of the tree. 
%Moreover, cluster sampling is conditioned on the tree state as well to avoid sampling of clusters too far from the tree or regions that were already exploited in the current run (see Sec. \ref{sec:adaptive-sampling} and \ref{sec:non-stationary}).

The next sections detail the clustering and sampling phases.
Tuning guidelines are given in Appendix \ref{appendix:tuning}.

\subsection{Online learning of high-reward regions}
\label{sec:clustering}

We aim to find groups of transitions that constitute high-reward regions.
Because full state-space coverage is often intractable, we do not try to create a partition of the entire space of possible transitions, instead focusing on the transitions obtained from previous iterations.
Given a set of transitions $\mathcal{T}$, we cluster them according to their reward and spatial distribution through the distance function:
\begin{equation}
\label{eq:clustering-metric}
    d(\tau_1, \tau_2) = || \tau_1.\x_{\parent} - \tau_2.\x_{\parent}|| + || \tau_1.\x_{\child} - \tau_2.\x_{\child}|| + \lambda || \rho(\tau_1) - \rho(\tau_2) ||
\end{equation}
where $\lambda > 0$, and $\rho(\tau_j)$ is the reward of $\tau_j$. 
Although any clustering techniques could be used, we use HDBSCAN because of its effectiveness at identifying irregular clusters and ease of tuning (see also Appendix \ref{appendix:tuning}).
Hence, we obtain a set of clusters $\{ \mathcal{C}_1, \dots , \mathcal{C}_N \}$ and each cluster is associated with an average reward $\mathcal{R}_i = \nicefrac{1}{|\mathcal{C}_i|} \sum_{\tau_j\in \mathcal{C}_i} \rho(\tau_j)$. 
These clusters are subsets of transitions, which will be used to bias sampling.

\begin{comment}
\begin{algorithm}[tpb]
\DontPrintSemicolon
\SetKwFunction{sampleCluster}{sampleCluster}
\SetKwFunction{selectNextBanditArm}{selectNextBanditArm}
\SetKwFunction{updateBanditArms}{updateBanditArms}
\SetKwFunction{nearest}{nearest}
\SetKwFunction{sampleTo}{sampleTo}
%
\KwIn{ $\{\mathcal{C}_i\}_{i=0}^n$, $\mathcal{G}$ }
%\KwOut{$x_{\parent}$, $x_{\target}$}
\KwOut{$x_{\parent}, \, x_{\target}$}
$x_{\parent} \leftarrow \CommentSty{NULL}$,\,\,$x_{\target} \leftarrow $  \CommentSty{NULL}\;
\While{$x_{\parent}=\CommentSty{NULL}$ \KwOr $x_{\target}=\CommentSty{NULL}$}
{
    $i \leftarrow $\selectNextBanditArm()\;
    \If{$i=1$}{ \label{alg:sample:uniform}
        $x_{\target} \leftarrow  \unif(\X)$\;
        $x_{\parent} \leftarrow \mathcal{G}.\nearest(x_{\target})$\;
    }
    \ElseIf{$i=2$}{
        $x_{\target} \leftarrow \unif(X_{\mathrm{goal}})$\;
        $x_{\parent} \leftarrow \mathcal{G}.\nearest(x_{\target})$\; \label{alg:sample:goal}
    }
    \Else{
        $x_{\parent},\, x_{\target} \leftarrow $ \sampleCluster($\mathcal{C}_i$, $\mathcal{G}$)\;
        \If{$x_{\parent}=\CommentSty{NULL}$ \KwOr $x_{\target}=\CommentSty{NULL}$}{
            $\updateBanditArms(0.0)$\;  \label{alg:sample:update0}
        }
    }
 }
\Return $\x_{\parent},\,  \x_{\target}$
\caption{\FuncSty{sample}}
\label{alg:sample}
\end{algorithm}
\end{comment}

\begin{algorithm}[tpb]
\DontPrintSemicolon
\SetKwFunction{sampleCluster}{sampleCluster}
\SetKwFunction{selectNextBanditArm}{selectNextBanditArm}
\SetKwFunction{updateBanditArms}{updateBanditArms}
\SetKwFunction{nearest}{nearest}
\SetKwFunction{sampleTo}{sampleTo}
\KwIn{ $\mathcal{C}$, $\mathcal{G}$ }
\KwOut{$(x_{\parent},\, u,\, d, \, x_{\child},\, x_{\target})$}
$x_{\parent} \leftarrow \CommentSty{NULL}$,\,
$x_{\target} \leftarrow $  \CommentSty{NULL}\;
\While{$x_{\parent}=\CommentSty{NULL}$ \KwOr $x_{\target}=\CommentSty{NULL}$}
{
    $i \leftarrow $\selectNextBanditArm()\; \label{alg:sample:next-arm}
    \If{$i=1$}{ \label{alg:sample:uniform}
        $x_{\target} \leftarrow  \unif(\X)$\;
        $x_{\parent} \leftarrow \mathcal{G}.\nearest(x_{\target})$\;
    }
    \ElseIf{$i=2$}{
        $x_{\target} \leftarrow \unif(X_{\mathrm{goal}})$\;
        $x_{\parent} \leftarrow \mathcal{G}.\nearest(x_{\target})$\; \label{alg:sample:goal}
    }
    \Else{
        $x_{\parent},\, x_{\target} \leftarrow $ \sampleCluster($\mathcal{C}_i$, $\mathcal{G}$)\;
        \If{$x_{\parent}=\CommentSty{NULL}$ \KwOr $x_{\target}=\CommentSty{NULL}$}{
            $\updateBanditArms(0)$\;  \label{alg:sample:update0}
        }
    }
 }
$x_\child,\,u,\,d \leftarrow \sampleTo(x_\parent, x_\target, i)$\; \label{alg:sample:sampleTo}
\Return $(x_{\parent},\, u,\, d, \, x_{\child},\, x_{\target})$
\caption{\FuncSty{sampleAndPropagate}}
\label{alg:sample}
\end{algorithm}

%\begin{algorithm}[tpb]
%\DontPrintSemicolon
%
%\SetKwFunction{sample}{sample}
%\SetKwFunction{propagate}{propagate}
%\SetKwFunction{add}{add}
%\SetKwFunction{updateReward}{updateReward}
%\SetKwFunction{findPath}{findPath}
%\SetKwFunction{clustering}{clustering}
%
%\KwIn{$\x_{\parent}$, $\x_{\target}$}
%\KwOut{$\x_{\mathrm{next}}$}
%\KwParameter{$K_u>0$}
%$\delta_{\mathrm{best}} \leftarrow +\infty$\;
%\For{$k$ \KwInterval $1,\dots,K_u$}
%{    
%    $u \leftarrow \unif(\U)$\;
%    $d \leftarrow \unif([d_{\mathrm{min}}, d_{\mathrm{max}}])$\;
%    $x_{\mathrm{prop}} \leftarrow \x_{\parent} + \int_0^d f(x(t), u) dt$\;
%    \If{$||x_\mathrm{prop} - x_\mathrm{prop}|| < \delta_{\mathrm{best}}$}{
%        $\delta_{\mathrm{best}} \leftarrow ||x_\target - x_\mathrm{prop}||$\;
%        $x_\mathrm{next} \leftarrow x_\mathrm{prop}$\;
%    }   
%    \Return $x_\mathrm{next}$\;
%}    
% \caption{\FuncSty{sampleTo}}
% \label{alg:sampleTo}
%\end{algorithm}

\subsection{Adaptive sampling of high-reward regions}
\label{sec:adaptive-sampling}

We bias the probability of sampling cluster $\mathcal{C}_i$ according to its expected reward.
We model the problem of finding the optimal sampling bias as an MAB problem, where each $\mathcal{C}_i$ is an arm and $\mathcal{R}_i$ is its initial reward.
We also consider uniform and goal sampling as arms of the MAB problem.
Therefore, we define a set of $M$ arms, $\{ a_1,\dots, a_M \}$, with $M=N+2$:
\begin{equation}
\label{eq:arms}
    a_i = \begin{dcases*}
        \text{uniform sampling over } \X & if $i=1$\\
        \text{uniform sampling over } X_{\mathrm{goal}} & if $i=2$\\
        \text{sampling } \mathcal{C}_{i-2} & if $i\geq3$
    \end{dcases*}
\end{equation}
MAB selects where to sample the next transition from (line \ref{alg:sample:next-arm} of Alg. \ref{alg:sample}).
If the MAB selects the first or second arm, $\tau.\x_{\target}$ is a random state or a goal state, respectively, while $\tau.\x_{\parent}$ is the node of the tree closest to $\tau.\x_{\target}$ (lines \ref{alg:sample:uniform}--\ref{alg:sample:goal} of Alg. \ref{alg:sample}).
For all other arms, the algorithm tries to sample a transition from the selected cluster (Alg. \ref{alg:sample-cluster}).
Cluster sampling depends on the search tree because:
\begin{itemize}
    \item[i] We want to avoid over-sampling regions that the current tree has already explored; thus, we discard a candidate $\tau_c$ if $\tau_c.\x_{\target}$ is too close to the current tree;
    \item[ii] We select $\tau_c$ only if its pre-conditions are met; \emph{i.e.}, if $\tau_c.\x_{\parent}$ is close enough to the current search tree;
    \item[iii] If it is impossible to sample a transition whose pre-conditions are met, we try to extend the tree in such a way as to meet the pre-conditions in future timesteps; \emph{i.e.}, we set $\tau_c.\x_{\parent}$ as the target of the new sample $\tau$.
\end{itemize}
Alg. \ref{alg:sample-cluster} implements these considerations. 
It randomly draws candidate transitions, $\tau_c$,  from a cluster and perturbs them until $\tau_c.\x_{\target}$ is further than $\delta_1$ from $\mathcal{G}$ and $\tau_c.\x_{\parent}$ is closer than $\delta_2$ to $\mathcal{G}$ (lines \ref{alg:sample-cluster:delta1}--\ref{alg:sample-cluster:delta2-end}).
If it is impossible to satisfy the second condition (\emph{i.e.}, $\mathcal{G}$ is too far from the selected cluster), we try to expand the tree toward that cluster by looking for a transition whose parent is closer than $\delta_3 > \delta_2$ to $\mathcal{G}$ (line \ref{alg:sample-cluster:delta3}) by selecting its parent as the new target.
If we could not draw a valid transition from the selected cluster, we discourage its future selection by assigning a reward equal to zero (line \ref{alg:sample:update0}).
Finally, \texttt{sampleTo} (line \ref{alg:sample:sampleTo} of Alg. \ref{alg:sample}) generates a transition by propagating a random action  for a random duration if $i
\leq 2$. If $i> 2$, \texttt{sampleTo} tries to extend toward $x_{\target}$ by sampling $N_p$  random controls, $\u\in \U$, and durations, $d\in[0,T_p]$, with $T_p>0$, and selecting the closest transition to $x_{\target}$.

\begin{algorithm}[tpb]
\DontPrintSemicolon
\SetKwFunction{sampleCluster}{sampleCluster}
\SetKwFunction{nearest}{nearest}
\KwIn{ $\mathcal{C}_i$, $\mathcal{G}$ }
\KwOut{$x_{\parent}$, $x_{\target}$}
\KwParameter{$K>0, \delta_1,  \delta_2 >0, \delta_3>\delta_2$, $w>0$}
    $x_{\parent} \leftarrow $ \CommentSty{NULL},\,\,
    $x_{\target} \leftarrow $  \CommentSty{NULL}\;
    \For{$k$ \KwInterval $1,...,K$}
    {
         $\tau_{\mathrm{c}} \leftarrow \unif(\mathrm{\mathcal{C}}_i)$\;
         $\tau_c.\x_{\parent} = \tau_c.\x_{\parent} + \unif([-w,w])$\; \label{alg:sample:perturbation-1}
         $\tau_c.\x_{\target} = \tau_c.\x_{\target} + \unif([-w,w])$\; \label{alg:sample:perturbation-2}
         \If{$|| \tau_c.{x_\target}\! - \mathcal{G}.\nearest(\tau_c.{x_\target})|| < \delta_1\!$ \label{alg:sample-cluster:delta1}} 
         {
            \KwContinue 
         }
         \If{$|| \tau_c.{x_\parent}\! - \mathcal{G}.\nearest(\tau_c.{x_\parent})|| \!<\! \delta_2\! $ \label{alg:sample-cluster:delta2}}{
            $x_{\parent} \leftarrow \mathcal{G}.\nearest(\tau_c.x_{\parent})$\;
            $x_{\target} \leftarrow \tau_c.{x_\target}$\;
            \KwBreak \label{alg:sample-cluster:delta2-end}
         }
         \If{$|| \tau_c.{x_\parent} \!- \mathcal{G}.\nearest(\tau_c.x_{\parent})|| \!< \!\delta_3\! $ \label{alg:sample-cluster:delta3}}{
             $x_{\parent} \leftarrow \mathcal{G}.\nearest(\tau_c.x_{\parent})$\;
            $x_{\target} \leftarrow \tau_c.{x_\parent}$\;
         }
     }
     \Return $x_{\parent}$, $x_{\target}$
 %}
 \caption{\FuncSty{sampleCluster}}
 \label{alg:sample-cluster}
\end{algorithm}

\subsection{Non-stationary rewards}  \label{sec:non-stationary}

We update the estimated reward distributions when we add a transition to the tree (line \ref{alg:MAB-RRT:updateMAB} of Alg. \ref{alg:MAB-RRT}) and if we could not sample a valid candidate transition from the selected cluster $\mathcal{C}_i$ (line \ref{alg:sample:update0} of Alg. \ref{alg:sample}).
In the first case, we use the reward $\rho(\tau)$ realized by the transition; in the second case, we use a reward equal to zero to discourage sampling $\mathcal{C}_i$ if it does not satisfy conditions i, ii, or iii from Sec. \ref{sec:adaptive-sampling}.
In both cases, the reward realized by a transition depends on the tree  state.
We model the variability of the reward with respect to the tree state as a non-stationary MAB problem, where the reward distribution can vary over iterations.
Standard MAB algorithms such as UCB-1 \cite{UCB1} perform poorly under these conditions because they adapt too slowly to the reward changes \cite{kfmanb}.
We therefore use a non-stationary bandit algorithm, which accounts for shifts in the reward distribution.
Specifically, we use the Kalman Filter-Based solution for Non-stationary Multi-Arm Bandit (KF-MANB) algorithm \cite{kfmanb}.
%, which uses Kalman filters to estimate the utility distribution of each arm and then uses Thompson sampling \cite{thompson} to select an arm at each timestep.
%The pseudo-code of KF-MANB is reproduced in Alg. \ref{alg:KF-MANB}.
KF-MANB models each arm's reward as a normal distribution.
At each iteration, it selects the next arm via Thompson Sampling \cite{thompson}.
%; that is, it draws a value from all arms' distributions and selects the arms that returned the higher value.
When it observes the reward, KF-MANB updates the estimated distributions using a Kalman Filter update rule, which allows for tracking non-stationary rewards over time.

\subsection{Completeness and optimality}
\label{sec:optimality}
\begin{comment}
\cite{Hauser:AO-RRT} proved that AO-RRT is asymptotically optimal if the underlying implementation of RRT is \emph{well-behaved}; \emph{i.e.}, it is probabilistically complete in the state-cost space.
In a refined version of the proof, \cite{bekris:refined-proof-aorrt} argues that this holds only for problems where (i) the dynamics system and the cost function are Lipschitz continuous, and (ii) the optimal trajectory is robust with clearance $\delta > 0$.
Note that not all variants of RRT are probabilistically complete \cite{stilman:rrt-not-complete}.
As shown in \cite{Bekris:proof-kinodynamic-rrt}, kinodynamic RRT is probabilistically complete under conditions (i) and (ii) if, at each iteration, the tree is extended by forward propagating random controls $\u \in \U$ for a random duration $d\in[0,T_p]$, with $T_p>0$.
This is true for our method
because \texttt{sampleTo}  in Alg. \ref{alg:sample} chooses random controls and durations when the first two arms (\emph{i.e.}, uniform and goal sampling) are chosen by the MAB algorithm.
Note that these arms have a non-zero probability of being selected at each iteration (like any other arm of the MAB).
Therefore, under the assumptions that $f(x,u)$ and $c(\sigma)$ from Sec. \ref{sec:preliminaries} are Lipschitz continuous and there exists a robust optimal solution with clearance $\delta>0$, our implementation of RRT is well-behaved and MAB-RRT is asymptotically optimal.
\end{comment}

Note that not all variants of RRT are probabilistically complete \cite{stilman:rrt-not-complete}.
Assumed that (i) the dynamics system is Lipschitz continuous, and (ii) there exists a robust solution with clearance $\delta > 0$,
kinodynamic RRT is probabilistically complete if it extends the tree by forward propagating random controls $\u \in \U$ for a random duration $d\in[0,T_p]$, with $T_p>0$ \cite{Bekris:proof-kinodynamic-rrt}.
This is true for our method because \texttt{sampleTo}  in Alg. \ref{alg:sample} chooses random controls and durations when the first two arms (\emph{i.e.}, uniform and goal sampling) are chosen by the MAB algorithm.
Note that these arms have a non-zero probability of being selected at each iteration (like any other arm of the MAB).
Therefore, under the assumptions that $f(x,u)$ from Sec. \ref{sec:preliminaries} is Lipschitz continuous and there exists a  solution with clearance $\delta>0$, each run of RRT in MAB-RRT is probabilistically complete.

As for asymptotic optimality, \cite{Hauser:AO-RRT} proved that AO-RRT is asymptotically optimal if the underlying RRT is \emph{well-behaved} in the augmented state-cost space.
In a refined version of the proof, \cite{bekris:refined-proof-aorrt} argues that well-behavedness holds if the dynamics system and the cost function derivative are Lipschitz continuous, and the optimal trajectory is robust with clearance $\delta > 0$, proving the asymptotic optimality of a single-tree version of AO-RRT.
Because Alg. \ref{alg:MAB-RRT} uses a multi-tree implementation and runs RRT in the state space, we cannot derive asymptotic optimality directly from \cite{bekris:refined-proof-aorrt}.
Nonetheless, results in Sec. \ref{sec:numerical} and \ref{sec:experiments} suggest that the solution cost decreases consistently with iterations.
We therefore leave the formal analysis of MAB-RRT asymptotic optimality as future work.

\section{Simulation Results}\label{sec:numerical}

This section shows that our approach improves the solution cost faster than AO-RRT and yields smaller cumulative regret with 2D problems. %
We consider a single integrator $\dot{x} = u$, $\X = [0,1]^2$, $\U\in[0.5,0.5]^2$, and the five scenarios in Fig. \ref{fig: 2d-cases}.
We consider the reward function $\rho(\tau) = 0.5(\rho_x(\tau.x_{\parent}) + \rho_x(\tau.x_{\child}))$ where $\rho_x\in [0,1]$ as in Fig. \ref{fig: 2d-cases} and the cost function $c(\sigma) = \sum_{\tau \in \sigma} (1-\rho(\tau)) ||\tau.x_{\parent} - \tau.x_{\child}||$.

The scenarios serve as illustrative examples of problems with different features.
For example, in Scenario A, the optimal solution should take a long path through a narrow passage, while Scenarios C and D are examples of ``trap'' problems, where the high-reward region leads to a dead end. Scenario  E combines these issues into a more complex problem.

\subsection{Cost analysis}\label{sec: cost-analysis}

We compare our MAB-RRT-KFMANB with AO-RRT \cite{Hauser:AO-RRT} and other variants of MAB-RRT using UCB-1 \cite{UCB1} and Thompson Sampling \cite{thompson}.
Fig. \ref{fig:planar:cost} shows the average cost trends for 30 repetitions.
Except for Scenario C, MAB-RRT-KFMANB  improves the solution significantly faster than AO-RRT, suggesting that the online bias learning drives the search to more promising regions.
In Scenario C, MAB-RRT-KFMANB  has a slightly worse convergence rate because the high-reward region (yellow in Fig. \ref{fig: 2d-cases}) leads to a dead end (notice that the optimal path mainly lies in the low-reward region).
MAB-RRT-KFMANB  outperforms AO-RRT even in Scenario D, where the high-reward region leads to a dead end: after an initial exploration of the high-reward, MAB-RRT spots the medium-reward region and quickly improve the solution cost.
As expected, MAB-RRT-KFMANB outperforms the stationary variants.
Overall, MAB-RRT-UCB1 and MAB-RRT-TS perform comparably to AO-RRT, showing the importance of the non-stationary MAB to account for the changing reward.

\subsection{Regret analysis} \label{sec: regret-analysis}

\begin{algorithm}[tpb]
\DontPrintSemicolon
%
%\SetKwFunction{sample}{sample}
%
\KwIn{tree $\mathcal{G}$, iteration $k$}
\KwOut{expected regret $E[R_{\mathrm{strategy}}]$ for all strategies }
$\emph{strategies}$ = \{ \!\!\!\! \texttt{kfmanb,\!\!\! ucb1,\!\!\! TS,\!\!\! random,\!\!\! astar} \!\!\}\;
\For{strategy \KwInterval strategies } { \label{alg:regret:batch-start}    
    \If{strategy \KwInterval \textup{\{\texttt{kfmanb,\!\!\! ucb1,\!\!\! TS\} }} }{
        draw a batch of points for each arm of the MAB and a batch of points using Alg. \ref{alg:sample}\;
    }
    \If{strategy = \textup{\texttt{random}} }{
        draw a batch of points using the \textup{\texttt{random}} sampling strategy\;
    }
    \If{strategy = \textup{\texttt{astar}} }{
        draw a batch of points using the \textup{\texttt{astar}} sampling strategy\;
    }

    For all batches of points,  generate the corresponding transitions with respect to $\mathcal{G}$;\; \label{alg:regret:batch-end}

}

For all batches of transitions, compute the expected reward ($\bar{r}_{\mathrm{rnd}}$, $\bar{r}_{A^*}$, $\bar{r}_{\texttt{kfmanb},i}$, $\bar{r}_{\texttt{ucb1},i}$, $\bar{r}_{\texttt{TS},i}$ $\forall i$=$1...M$)\; \label{alg:regret:rewards}

Compute the best expected reward $E[r^*(k)]$ as the maximum of the average rewards of all batches;\; \label{alg:regret:best-arm}

\For{strategy \KwInterval strategies }
{
regret $E[R_{\mathrm{strategy}}] = E[r^*(k)] - \bar{r}_{\mathrm{strategy}}$;\; \label{alg:regret:regret}
}
 \caption{Regret computation}
 \label{alg:regret}
\end{algorithm}

\begin{figure*}[tbp]
	\centering
    % grid,tics=5
    \hspace{-1cm}
    \begin{overpic}[trim = 0.1cm 0cm 0cm 0.6cm, clip, angle=0, width=0.75\textwidth]{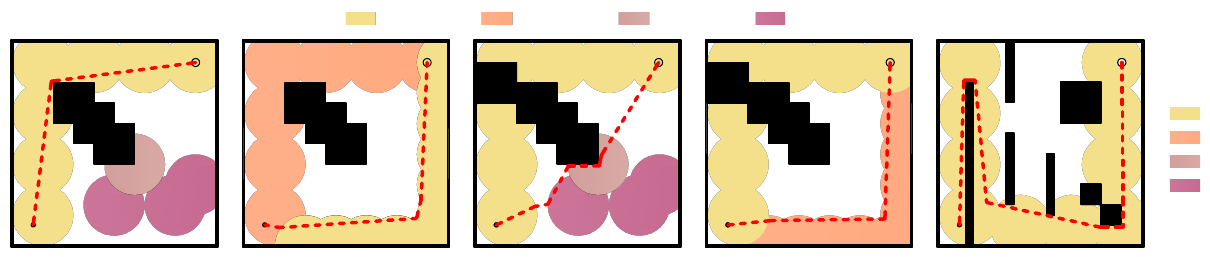} 
    \put (3,1.6) {\footnotesize$\x_{\mathrm{start}}$}
    \put (11,14) {\footnotesize$X_{\mathrm{goal}}$}
    \put (20.5,4) {\footnotesize$\x_{\mathrm{start}}$}
    \put (29,15) {\footnotesize$X_{\mathrm{goal}}$}
    \put (42,2) {\footnotesize$\x_{\mathrm{start}}$}
    \put (48,15) {\footnotesize$X_{\mathrm{goal}}$}
    \put (60,4) {\footnotesize$\x_{\mathrm{start}}$}
    \put (67,15) {\footnotesize$X_{\mathrm{goal}}$}
    %\put (73.25,2) {\footnotesize$\x_{\mathrm{start}}$}
    \put (86.5,16) {\footnotesize$X_{\mathrm{goal}}$}
    % rewards legend
    \put (100,11.5) {\footnotesize $\rho_x = 0.99$}
    \put (100,9.5) {\footnotesize $\rho_x = 0.8$}
    \put (100,7.5) {\footnotesize $\rho_x = 0.2$}
    \put (100,5.5) {\footnotesize $\rho_x = 0.1$}
    % captions
    \put (5,-2) {\footnotesize Scenario A}
    \put (24,-2) {\footnotesize Scenario B}
    \put (43,-2) {\footnotesize Scenario C}
    \put (62,-2) {\footnotesize Scenario D}
    \put (81,-2) {\footnotesize Scenario E}
    \end{overpic}
    \vspace{0.3cm}
	\caption{Scenarios used for the numerical analysis in Sec. \ref{sec:numerical}. Dashed red lines are the optimal paths.}
	\label{fig: 2d-cases}
    \vspace{-0.5cm}
\end{figure*}
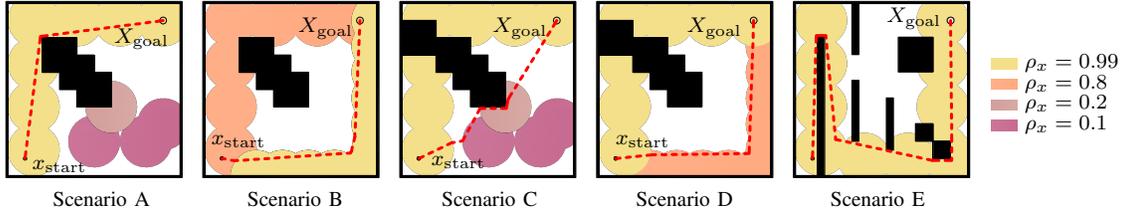

\begin{figure*}[tpb]
	\centering
	\subfloat[][Scenario A]
	{\includegraphics[trim = 0cm 0cm 0.0cm 0cm, clip, angle=0, height=0.285\columnwidth]{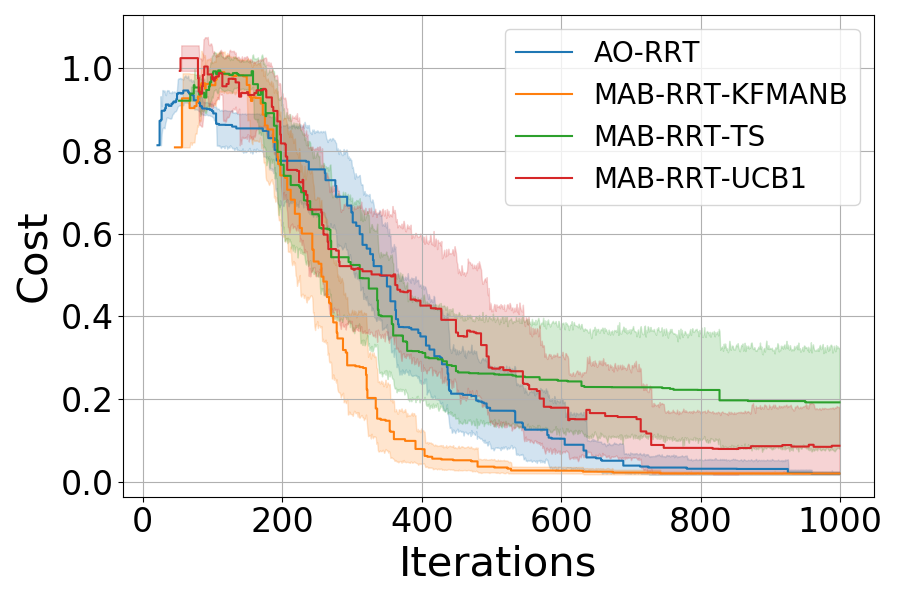}\label{fig:planar:cost:A}}
	\subfloat[][Scenario B]
	{\includegraphics[trim = 1.9cm 0cm 0cm 0cm, clip, angle=0, height=0.285\columnwidth]{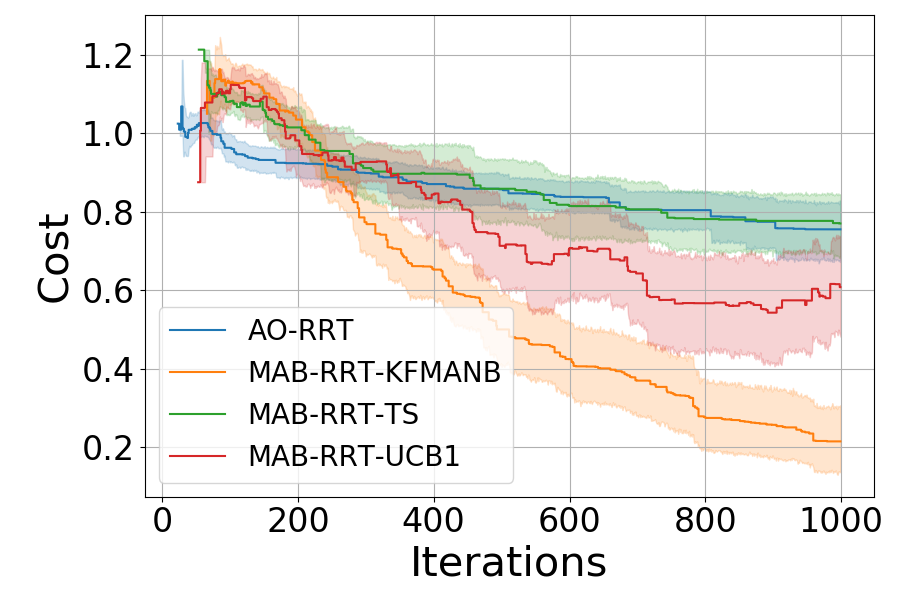}}
	\subfloat[][Scenario C]
	{\includegraphics[trim = 1.5cm 0cm 0cm 0cm, clip, angle=0, height=0.285\columnwidth]{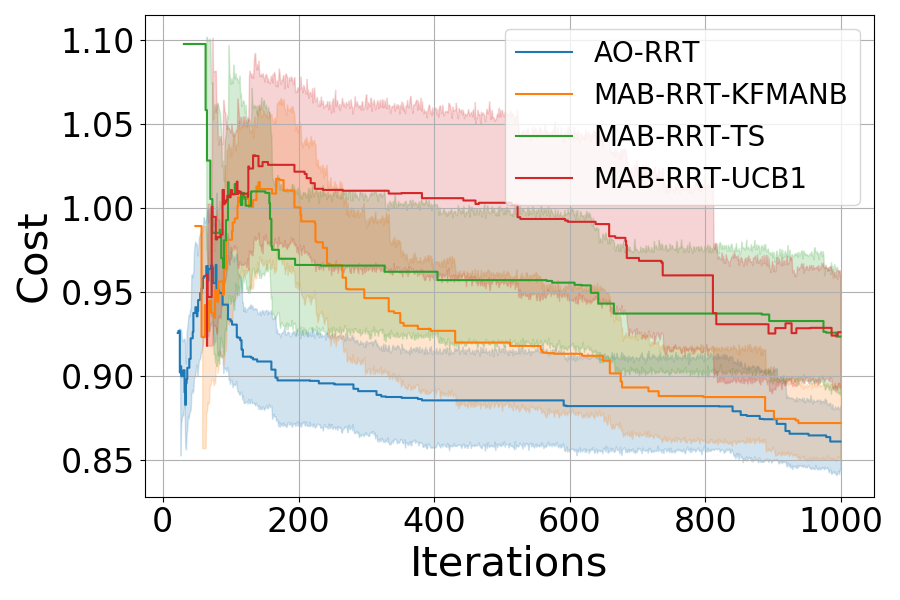}\label{fig:planar:cost:C}}
    \subfloat[][Scenario D]
	{\includegraphics[trim = 1.5cm 0cm 0cm 0cm, clip, angle=0, height=0.285\columnwidth]{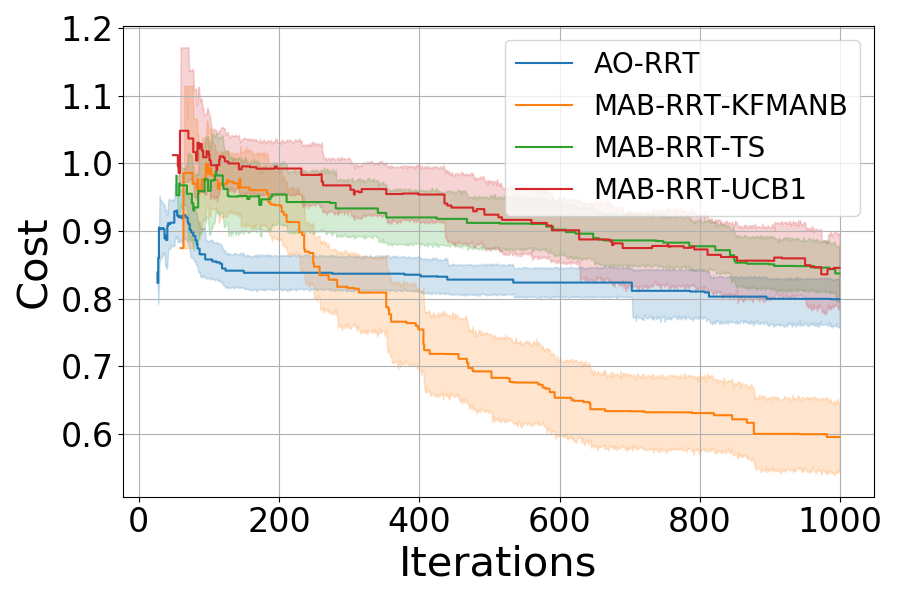}
    \label{fig:planar:cost:D}}
    \subfloat[][Scenario E]
	{\includegraphics[trim = 1.5cm 0cm 0cm 0cm, clip, angle=0, height=0.285\columnwidth]{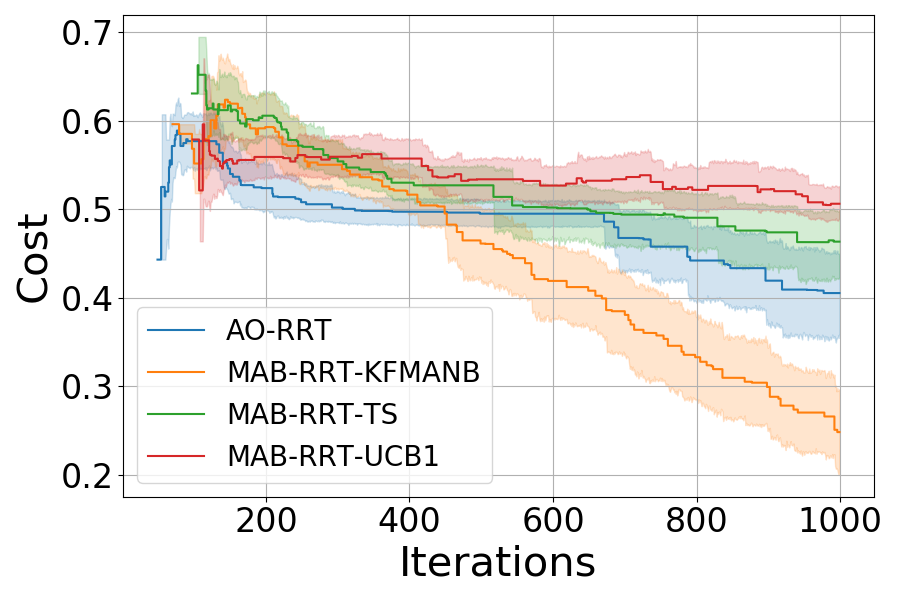}
    \label{fig:planar:cost:E}}
	\caption{Cost trends for the scenarios of Fig. \ref{fig: 2d-cases} (30 repetitions; solid lines: mean values; shadow: 95\% confidence interval).}
\label{fig:planar:cost}
\vspace{-0.5cm}
\end{figure*}

A standard metric to evaluate MAB is regret, \emph{i.e.}, the difference between the reward one would have obtained by sampling the best action and the reward realized by the chosen action at iteration $k$.
We can define the expected regret of a sampling strategy over a (possibly infinite) horizon $K$ as 
\begin{equation}
    E[R(K)] = \sum_{k=1}^K E[r^*_k]  -  E[r_k]
\end{equation}
where $r^*_k$ and $r_k$ are the rewards of the best sampling strategy and the chosen sampling strategy at iteration $k$, respectively.
We evaluate the regret of the following sampling strategies:
\begin{itemize}
    \item \texttt{kf-manb}, \texttt{ucb1}, \texttt{TS}: our method as described in Alg. \ref{alg:sample} using KF-MANB, UCB-1, and Thompson Sampling;
    \item \texttt{random}: uniform sampling over $\X$, as in standard RRT;
    \item \texttt{astar}: it mimics the A$^*$ search strategy; the control space is discretized as $\bar{U} = \{-0.5,-0.25,0,0.25,0.5\}^2$ and, at each iteration, the node with the lower estimated cost and with unexplored children is expanded. We use $h(x_1,x_2)=(1-\max_x \rho_x(x) )|| x_1 -x_2 ||$ as admissible heuristic, where $\max_x\rho_x(x) = 0.99$ according to Fig. \ref{fig: 2d-cases}.
\end{itemize}
%We compute the best expected reward as
%\begin{multline}
%\label{eq:best-arm}
%    E[r^*(k)] = \max \big( 
%    \max_i \bar{r}_{\texttt{kfmanb},i}, 
%    \max_i \bar{r}_{\texttt{ucb1},i}, \\
%    \max_i \bar{r}_{\texttt{TS},i},  
%    \bar{r}_{\mathrm{rnd}}, 
%    \bar{r}_{A^*}  \big)
%\end{multline}
%where $r_{\mathrm{rnd}}$ and $r_{A^*}$ are the rewards obtained by \texttt{random} and \texttt{astar}, \rev{and $E[r_i]$ is the average reward of a batch of transitions drawn from the $i$th arm}.
The regret computation is described in Alg.  \ref{alg:regret}, which runs at the beginning of each iteration $k$ in Alg. \ref{alg:MAB-RRT}.
It samples a batch of points from each arm of each sampling strategy (lines \ref{alg:regret:batch-start}--\ref{alg:regret:batch-end}) and computes the average reward of each one (line \ref{alg:regret:rewards}).
The best arm reward is the maximum average reward across all batches (line \ref{alg:regret:best-arm}) and is used to compute the regret of each strategy (line \ref{alg:regret:regret}).
Note that the regret comparison requires the sampling strategies to be evaluated at each iteration given the same tree.
For this reason, we grow the tree by using samples from \texttt{kfmanb}  to obtain the tree at the next timestep.

\begin{figure*}[tpb]
	\centering
	\subfloat[][Scenario A]
	{\includegraphics[trim = 0cm 0cm 0cm 14.5cm, clip, angle=0, height=0.3\columnwidth]{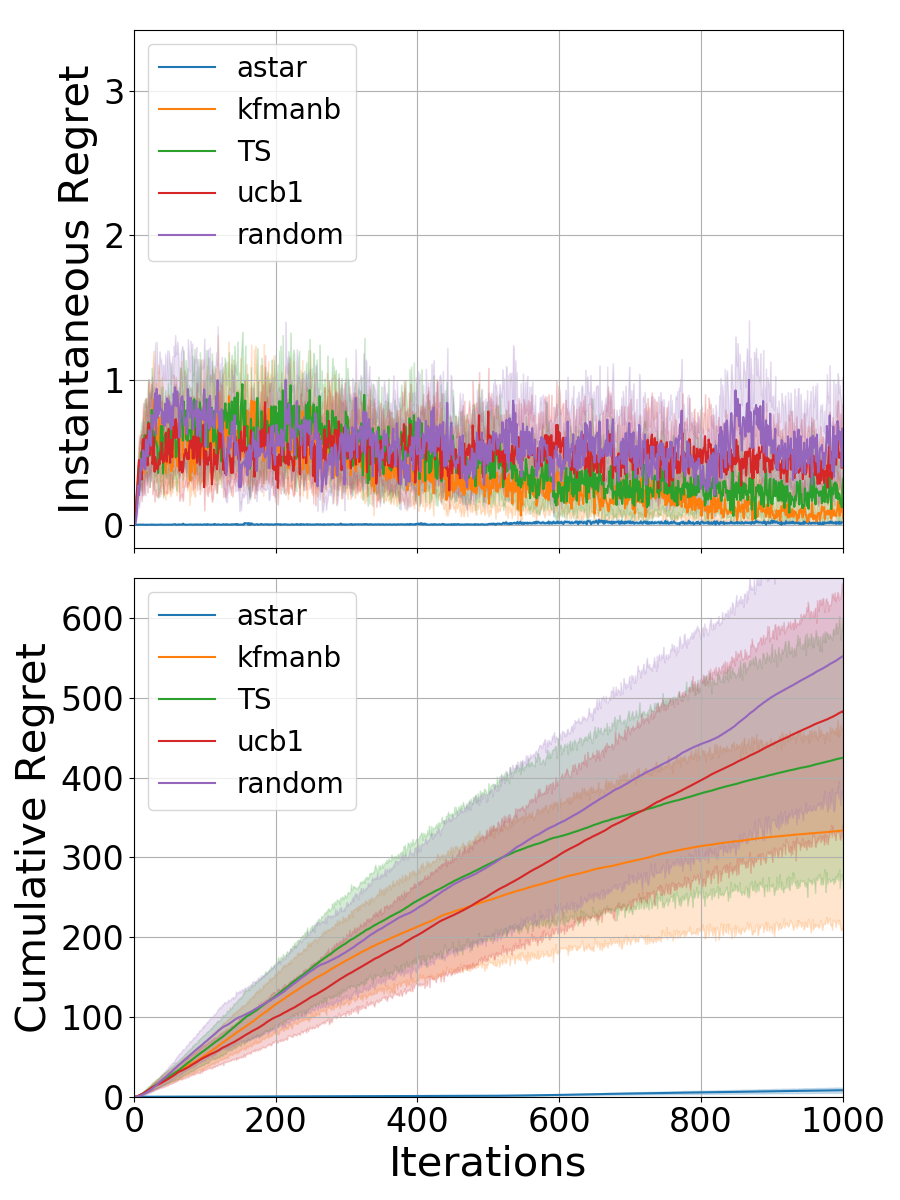}\label{fig:planar:regret-mab:A}}
	\subfloat[][Scenario B]
	{\includegraphics[trim = 1.7cm 0cm 0cm 14.5cm, clip, angle=0, height=0.3\columnwidth]{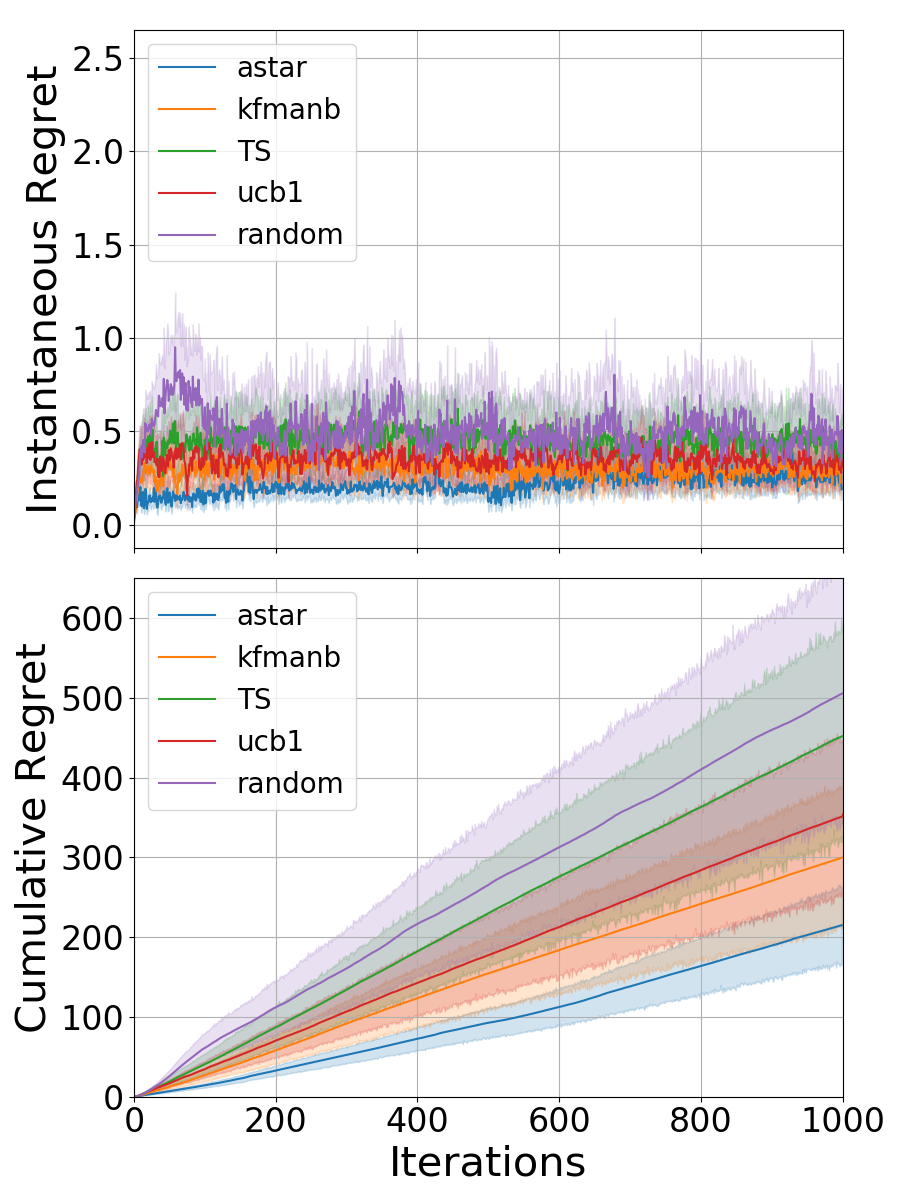}\label{fig:planar:regret-mab:B}}
	\subfloat[][Scenario C]
	{\includegraphics[trim = 1.5cm 0cm 0cm 14.5cm, clip, angle=0, height=0.3\columnwidth]{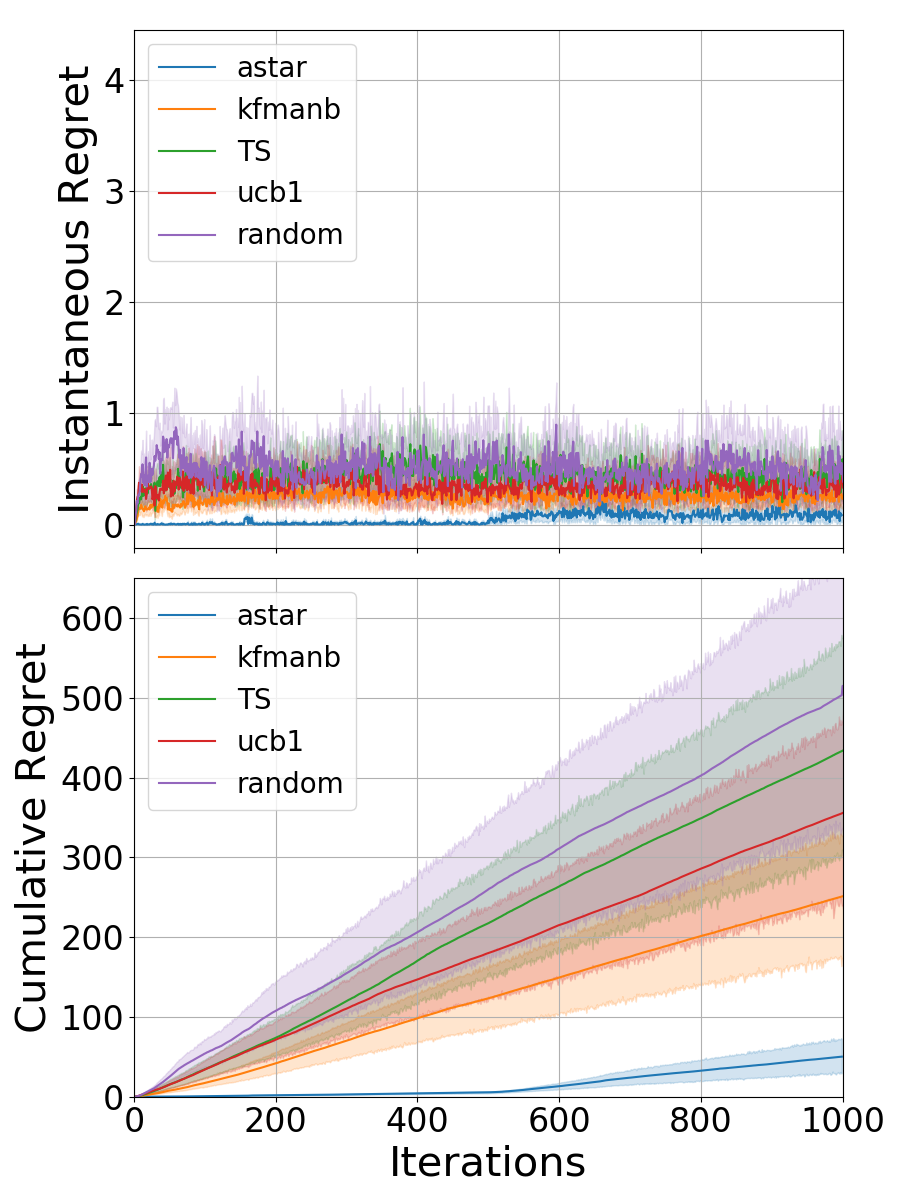}\label{fig:planar:regret-mab:C}}
    \subfloat[][Scenario D]
	{\includegraphics[trim = 1.5cm 0cm 0cm 14.5cm, clip, angle=0, height=0.3\columnwidth]{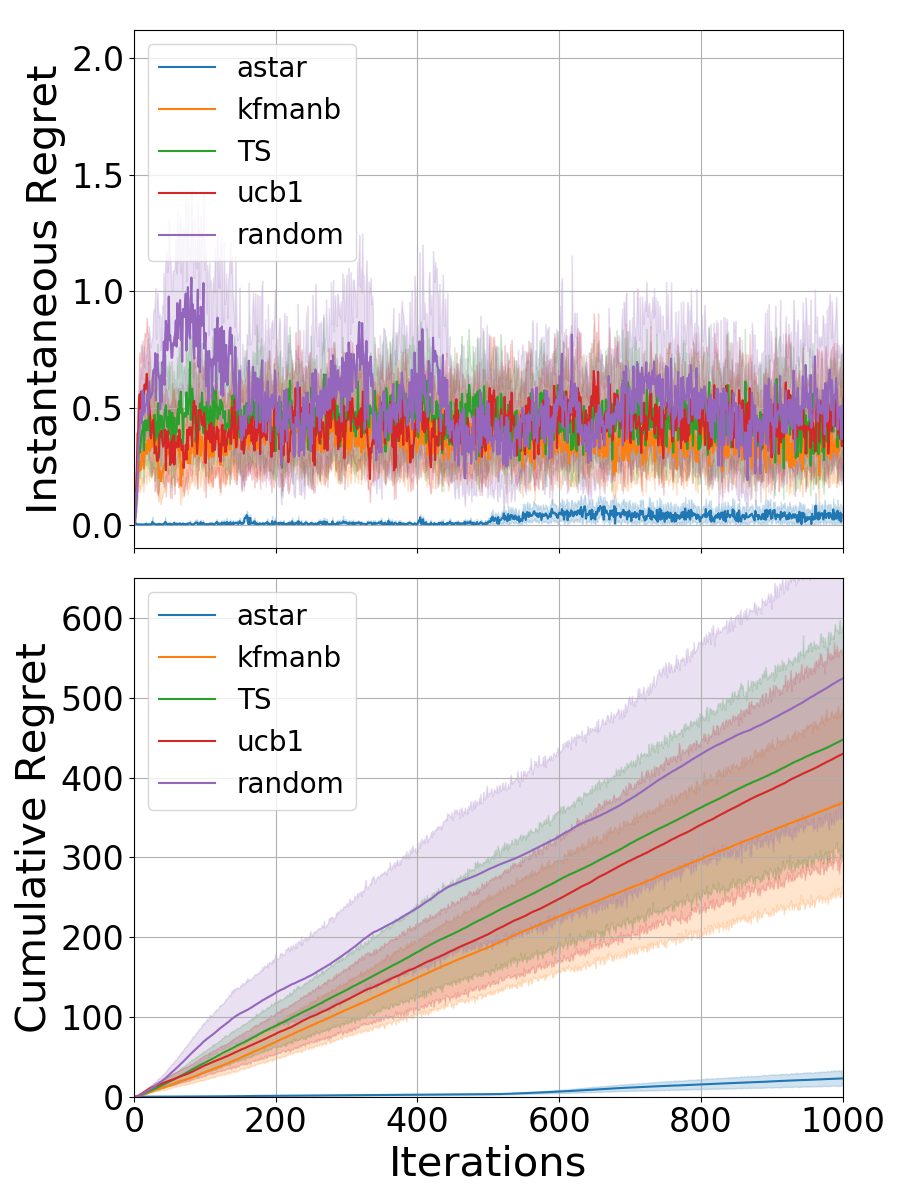}\label{fig:planar:regret-mab:D}}
    \subfloat[][Scenario E]
	{\includegraphics[trim = 1.5cm 0cm 0cm 14.5cm, clip, angle=0, height=0.3\columnwidth]{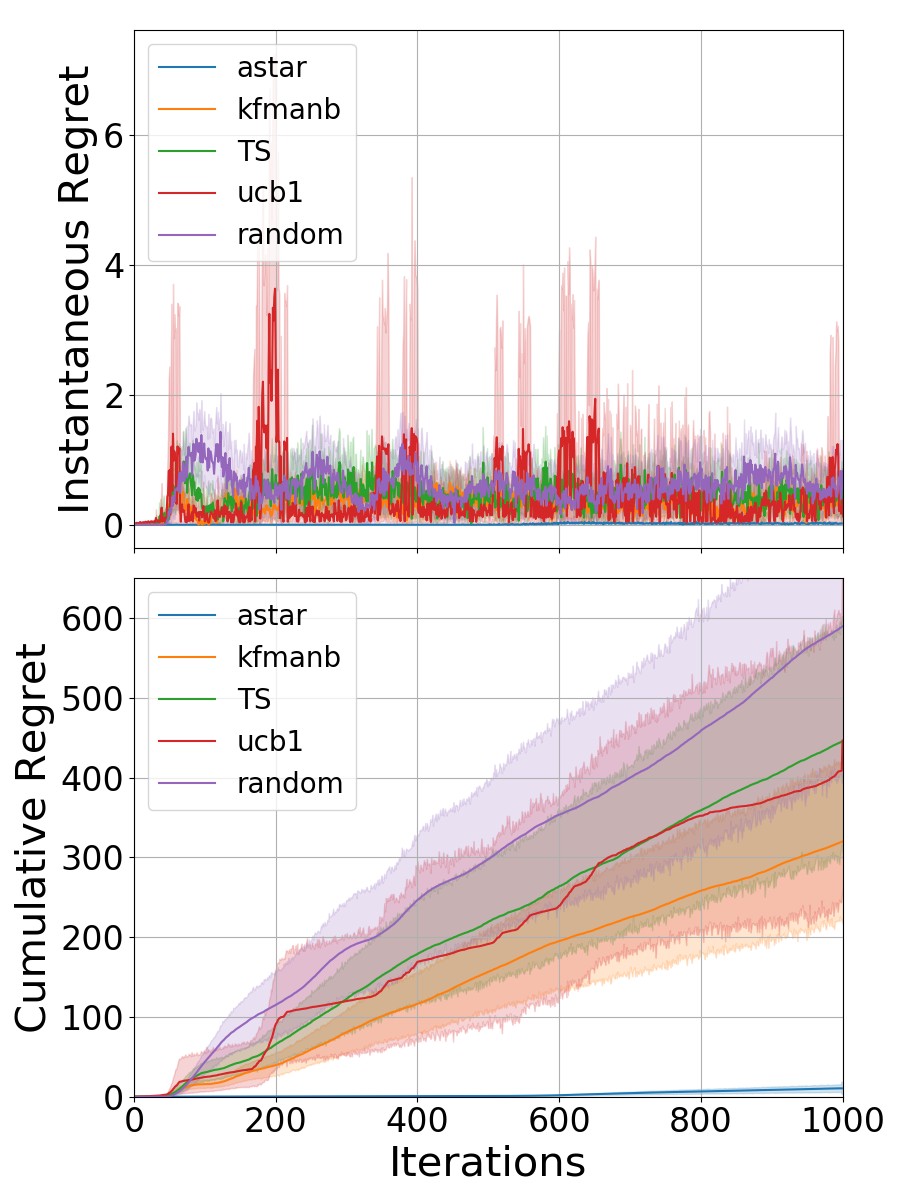}\label{fig:planar:regret-mab:E}}
 \caption{Cumulative regret for the scenarios of Fig. \ref{fig: 2d-cases} 
 (mean values of 30 repetitions).
 %(30 repetitions; solid lines: mean values; shadow: 95\% confidence interval).
 }
\label{fig:planar:regret-mab}
\vspace{-0.4cm}
\end{figure*}

Results are in Fig. \ref{fig:planar:regret-mab}.
Interestingly, \texttt{astar} yields very small regret in four out of five scenarios; 
\texttt{kfmanb} yields significantly lower regret than \texttt{random}, \texttt{ucb1} and \texttt{TS},  suggesting a correlation between the cost and the regret for all scenarios.

\begin{comment}
\begin{figure*}[tbp]
	\centering
    % grid,tics=5
    \begin{overpic}[trim = 0.1cm 0cm 1.1cm 0cm, clip, angle=0, width=0.8\textwidth]{img/mazes cases.pdf} 
    \put (1.5,2) {$\x_{\mathrm{start}}$}
    \put (14.5,20.5) {$X_{\mathrm{goal}}$}
    \put (26.75,2) {$\x_{\mathrm{start}}$}
    \put (39.85,20.5) {$X_{\mathrm{goal}}$}
    \put (52,2) {$\x_{\mathrm{start}}$}
    \put (65.1,20.5) {$X_{\mathrm{goal}}$}
    \put (77.25,2) {$\x_{\mathrm{start}}$}
    \put (90.3,20.5) {$X_{\mathrm{goal}}$}
    \put (30,25.3) {\footnotesize $\rho_x = 0.99$}
    \put (45,25.3) {\footnotesize $\rho_x = 0.8$}
    \put (59.5,25.3) {\footnotesize $\rho_x = 0.2$}
    \put (74.25,25.3) {\footnotesize $\rho_x = 0.1$}
    % captions
    \put (8.6,-2) {\footnotesize Scenario A}
    \put (33.85,-2) {\footnotesize Scenario B}
    \put (59.1,-2) {\footnotesize Scenario C}
    \put (84.35,-2) {\footnotesize Scenario D}
    \end{overpic}
    \vspace{0.3cm}
	%\includegraphics[trim = 0.1cm 0cm 1.1cm 0cm, clip, angle=0, width=\textwidth]{img/mazes cases.pdf} 
	\caption{Scenarios used for the numerical analysis in Sec. \ref{sec:numerical}. \TODO{print optimal paths}}
	\label{fig: 2d-cases}
    \vspace{-0.3cm}
\end{figure*}
\end{comment}

\subsection{Discussion}

The results confirm that the proposed sampling technique reduces the cumulative regret compared to uniform sampling.
The regret reduction coincides with smaller trajectory costs compared to AO-RRT.
Interestingly, A$^*$ accumulates the lowest regret because its heuristic search tends to expand nodes with low cost-to-come, focusing more on high-reward regions.
Unfortunately, A$^*$ is inefficient for high-dimensional problems, making the proposed method attractive from the perspective of high-dimensional kinodynamic planning.

Computationally, MAB-RRT differs from AO-RRT in the MAB-based sampling and the clustering. 
Given the same number of control propagations, $N_p$, in the \texttt{sampleTo} function, AO-RRT and MAB-RRT perform the same number of collision checks and forward dynamics propagations, leading to similar average iteration time (0.26 ms and 0.29 ms, respectively, with $N_p$ = $100$). 
%The MAB algorithm is computationally negligible, while \texttt{sampleCluster} requires additional nearest neighbor computations to check the suitability of the sampled transition. 
Clustering time grows with the number of transitions and the state-space dimensionality.
With an off-the-shelf Python implementation of HDBSCAN \cite{hdbscan-python} we observed an almost-linear clustering time between 2 and 80 ms for 100 and 5000 transitions.
Because of clustering and computational overheads, the average total planning time was 0.60 s for MAB-RRT and 0.43 s for AO-RRT.
Note that such difference is expected to become thinner for more complex scenarios, where the iteration time is predominant compared to the clustering time.

\section{Experiments}\label{sec:experiments}

We demonstrate our approach on a manipulation task with uncertain dynamics and show that it finds trajectories with a lower cost and a higher success rate in execution.
%We test the proposed method on a 7-degree-of-freedom Kuka IIWA 7 arm.
We consider a tabletop application where a Kuka IIWA 7 arm moves a dumbbell (6 kg) across the table, as in Fig.
\ref{fig: victor-dumbbell}.
Note that the payload (gripper + dumbbell) is around 9 kg, exceeding the maximum payload of the robot.
We implemented all the planners with OMPL \cite{OMPL} in a ROS/Gazebo \cite{gazebo} simulation environment.
The robot is controlled in joint-space impedance mode, which makes it compliant with the environment, yet causes large tracking errors with large payloads.
The state and control spaces are joint position and velocity, respectively.
We devise a cost function proportional to the robot Cartesian-space tracking error and dependent on the robot joint states:
\begin{equation}
\label{eq:kuka-cost-function}
    c(\sigma) = \sum_{\tau \in \sigma} \hat{e}(\tau)  ||\tau.x_{\parent} - \tau.x_{\child}|| \text{ and }
%\label{eq:kuka-reward-function}
    \rho(\tau) = 1 - \hat{e}(\tau) 
%    r(\tau) = 0.5\big(\, \hat{e}(\tau.x_{\parent}) + \hat{e}(\tau.x_{\child})\, \big)
\end{equation}
where $\hat{e} \in [0,1]$ is proportional to the end-effector Cartesian-space error (see Appendix \ref{appendix} for its derivation).
By minimizing $c$, the planner is expected to find trajectories that avoid large tracking errors, thus reducing the risk of unexpected collisions.

We compare MAB-RRT (with KF-MANB) and AO-RRT over three queries repeated 30 times.
Fig. \ref{fig: victor-results: cost} and \ref{fig: victor-results: success} show the path cost and the execution success rate on the real robot (\emph{i.e.}, the percentage of runs that reached the goal without collisions). 
To compare different queries, costs were normalized with respect to the best cost found for the corresponding query.
Fig. \ref{fig: victor-results: cost} shows that MAB-RRT's cost after 1500 iterations is significantly smaller than that of AO-RRT (-25\%).
Intuitively, a good solution avoids configurations where the payload causes a large deviation from the path.
This translates into a higher success rate when the trajectory is executed in the real-world (+65\%).
%This is shown in the video attached to the manuscript 
%The top images refer to the execution of a path computed by MAB-RRT, showing that the robot avoids the red obstacle by retracting the arm.
As shown in Fig. \ref{fig: victor-video-frames} and in the attached video, the paths found by MAB-RRT are more likely to avoid obstacles by retracting the arm (top images).
On the contrary, the path computed by AO-RRT (bottom images) passes above the obstacle while stretching the arm. In this configuration, the high payload causes a large path deviation, resulting in an unexpected collision.
Fig. \ref{fig: victor-results: regret} also shows the cumulative regret of \texttt{kfmanb} and \texttt{random} for a single experiment.
The trend qualitatively confirms the regret results discussed in Sec. \ref{sec: regret-analysis}.

\begin{comment}   
\begin{table}[tpb]
    \caption{
    Experimental results for the 7-degree-of-freedom manipulation task comparing AO-RRT and MAB-RRT.
	}	
	\label{tab: victor-results}
	\centering
	$
	\begin{array}{lcc}
	\toprule
	& \text{MAB-RRT} & \text{AO-RRT} \\
	& \text{mean (std.dev.)} & \text{mean (std.dev.)} \\
	\midrule
    \text{Path cost}  & \textbf{1.66}(4.5)   & 2.21(0.9) \\
    \text{Execution success rate}  & \textbf{0.95} & 0.58 \\
	\bottomrule
	\end{array}
	$
\end{table}
\end{comment}

\section{Conclusions}\label{sec:conclusions}
We presented an online learning approach to biased sampling for kinodynamic motion planning.
The approach runs RRT multiple times and uses MAB to choose between uniform sampling and sampling regions identified during the previous runs.
We showed that the proposed approach finds better solutions faster than an unbiased planner.
The experiments also suggested a correlation between low regret and cost in different scenarios.
Future works will investigate how to improve the performance of the approach, e.g., via pruning and a single-tree implementation as in \cite{bekris:refined-proof-aorrt} and  \cite{bekris:sst}.

\appendix

\subsection{Parameter tuning} \label{appendix:tuning}

We tune the parameters of MAB-RRT planner almost independently for each main module of the method.

\subsubsection{Clustering}
HDBSCAN requires the minimum number $N_{\mathrm{min}}$ of points in a cluster. 
Small values of $N_{\mathrm{min}}$ favor the identification of small clusters with sparse data (which is likely the case for high-dimensional planning problems).
Because small values of $N_{\mathrm{min}}$ allow for spotting small high-reward clusters, we empirically set $N_{\mathrm{min}}$ at random between 2 and 5 at each clustering.
Moreover, the reward weight $\lambda$ in \eqref{eq:clustering-metric} is needed to define the clustering distance function.
Large values of $\lambda$ tend to favor clusters with similar rewards; small values favor state-space proximity.
We observed a low sensitivity to $\lambda$ in all our experiments; $\lambda \in [1,10]$ yielded satisfactory performance.

\subsubsection{Bandit algorithm}

KF-MANB requires initializing the expected rewards $\bar{\mu}_i(0)$, their variance $\sigma_i^2(0)$, and the Kalman filter's noise factors $\sigma_{\mathrm{obs}}^2$, $\sigma_{\mathrm{tr}}^2$ and $\eta$.
$\sigma^2_{\mathrm{obs}}$ controls how much we believe the new observed reward (the larger $\sigma^2_{\mathrm{obs}}$, the faster the Kalman Filter adapts the arm's distribution mean). $\sigma_{\mathrm{tr}}^2$ increases the variance of non-selected arms to favor their exploration. $\eta$ is a tuning parameter to scale the covariance to match the reward scale.
While performance on individual experiments could be marginally improved by using different values, we found that $\sigma_i(0) = 0.2 \, \forall i\in1,\dots,M$, $\sigma_{\mathrm{obs}}^2 = 10^{-4}$ and $\sigma_{\mathrm{tr}}^2=10^{-4}$ gave satisfactory results for all of our planning tasks.
Concerning the initial rewards, we set $\bar{\mu}_i(0) = \mathcal{R}_i \, \forall i\in1,\dots,M$.
Finally, we set $\eta$ dynamically to 
$\eta(k+1) = \max(10^{-10}, 0.9\eta(k) + 0.1 |r(k+1)|)$ as in \cite{Berenson:estimafte-model-utility}.

\subsubsection{Cluster sampling}
Alg. \ref{alg:sample-cluster} requires thresholds $\delta_1$, $\delta_2$, and $\delta_3$.
%, which are respectively the distance for rejecting a target sample if too close to an existing node, the maximum distance for considering preconditions met for a sampled transition, and the maximum  distance for trying to create preconditions for a sampled transition.
We relate their values to the dispersion of each cluster so that, for all clusters,  $ \delta_1 = \delta_2 = \mathrm{median}(\{d_1,...,d_{|\mathcal{T}_i|}\})$, where
$ d_j = \min_z || z - \tau  || \, \forall \tau \in \mathcal{T}_i  \}$, and $\delta_3 = 2\, \delta_2$.

\subsection{Cost function proportional to Cartesian error} \label{appendix}

%To derive a function proportional to the maximum Cartesian-space tracking error, we consider the robot inverse dynamics 
%$
%    \xi = H(q) \ddot{q} + C(q,\dot{q})\dot{q} + g(q) + J(q)^T f_{\mathrm{ext}}
%$
%where $\xi,\,q,\,\dot{q} \in \mathbb R^{7}$ are the joint torque, position, and velocity, $H \in \mathbb R^{7\times 7}$ is the intertia matrix, $C \in \mathbb R^{7\times 7}$ accounts for Coriolis and centrifugal torques, $g \in \mathbb R^{7}$ is the gravity torque vector, $J \in \mathbb R^{6\times 7}$ is the robot Jacobian, and $f_{\mathrm{ext}} \in \mathbb R^{6}$ is the vector of external wrenches (owed to the payload).
Assuming we do not know the actual controller parameters, we consider a simplified proportional joint-space controller
$
    \xi_{\mathrm{mot}} = \hat{H}(q) K_p e_p + \hat{C}(q,\dot{q})\dot{q} + \hat{g}(q),
$
where $\xi_{\mathrm{mot}} \in \mathbb R^7$ is the torque required to the joint motors, $K_p>0$, and $\hat{H}$, $\hat{C}$, $\hat{g}$ are the estimated inertia, Coriolis, and gravity matrices. 
Assuming quasi-static conditions and perfect knowledge of robot inverse dynamics, we can write
$
       \hat{H}(q) K_p e_p - J(q)^T f_{\mathrm{ext}} = 0,
$
where $f_{\mathrm{ext}} \in \mathbb R^{6}$ is the external wrench (owed to the payload), and $J$ is the robot Jacobian.
By approximating $\Delta x \approx J \Delta q$ for small $\Delta q$, the estimated maximum Cartesian-space position error is
%\label{eq:error-kuka}
$
    e_{\mathrm{xyz}} = K_p^{-1}\, [I_3 \, 0_{3\times 3}]\, J(q) H(q) J(q)^T f_{\mathrm{ext}}.
$
Because $K_p$ is a constant scalar, we can set $K_p=1$ and scale $e_{\mathrm{xyz}}$ between $0$ and $1$ to obtain
$
     \hat{e}_q= \min\left(\nicefrac{\max(|e_{\mathrm{xyz}}|)}{e_{\mathrm{max}}}, 1\right) 
$
and $\hat{e}(\tau) = 0.5(\hat{e}_q(\tau.\x_{\parent}) + \hat{e}_q(\tau.\x_{\child}))$,
where $e_{\mathrm{max}}$ is an empirical estimate of the maximum value of $\max(|e_{\mathrm{xyz}}|)$.
In our experiments, we set $e_{\mathrm{max}}=70$ by computing the maximum value of $\max(|e_{\mathrm{xyz}}|)$ from $10^5$ random $q$.
$\hat{H}$ and $J$ are from the URDF model provided by the robot manufacturer.

\begin{figure}[tpb]
	\centering
 %\vspace{-0.2cm}
	\subfloat[][Path cost]
    {\includegraphics[trim = 0cm 0cm 0cm 0cm, clip, angle=0, height=0.37\columnwidth]{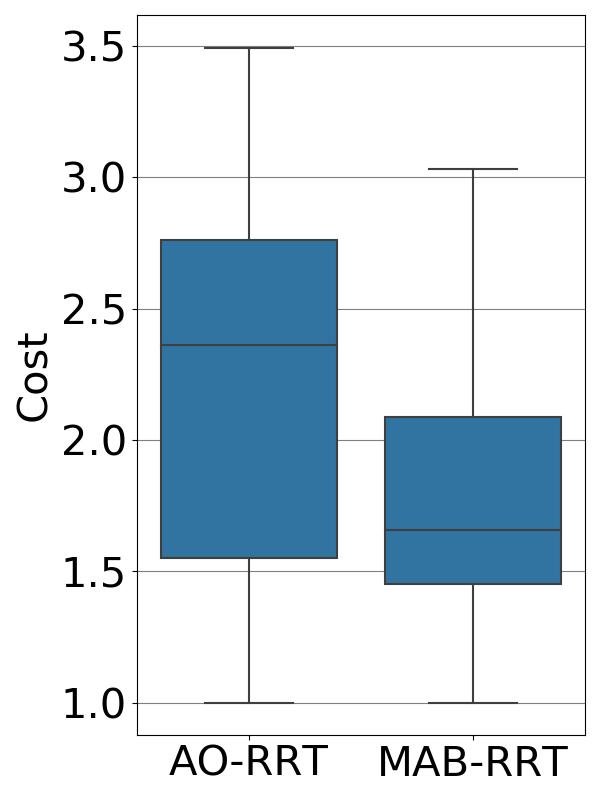}\label{fig: victor-results: cost}}\,\,
    \subfloat[][Execution success]
    {\includegraphics[trim = 0cm 0cm 0cm 0cm, clip, angle=0, height=0.37\columnwidth]{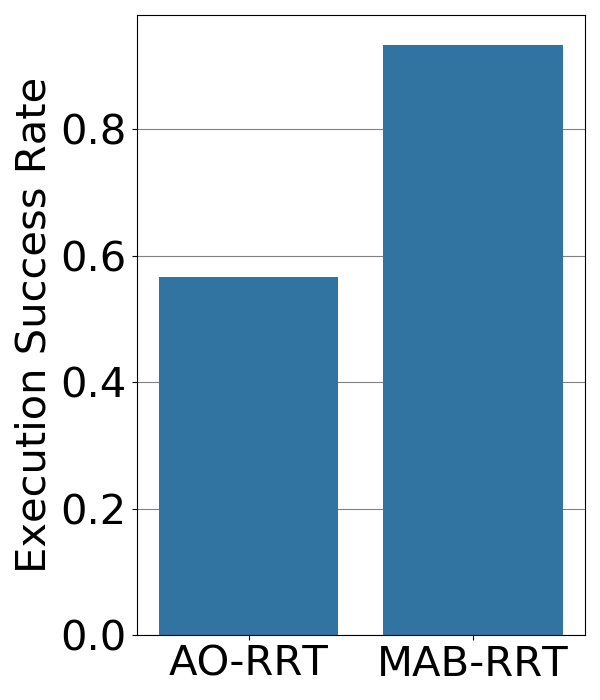}\label{fig: victor-results: success}}\,\,
	\subfloat[][Cumulative regret]
    {\includegraphics[trim = 0.0cm 0cm 0cm 18cm, clip, angle=0, height=0.37\columnwidth]{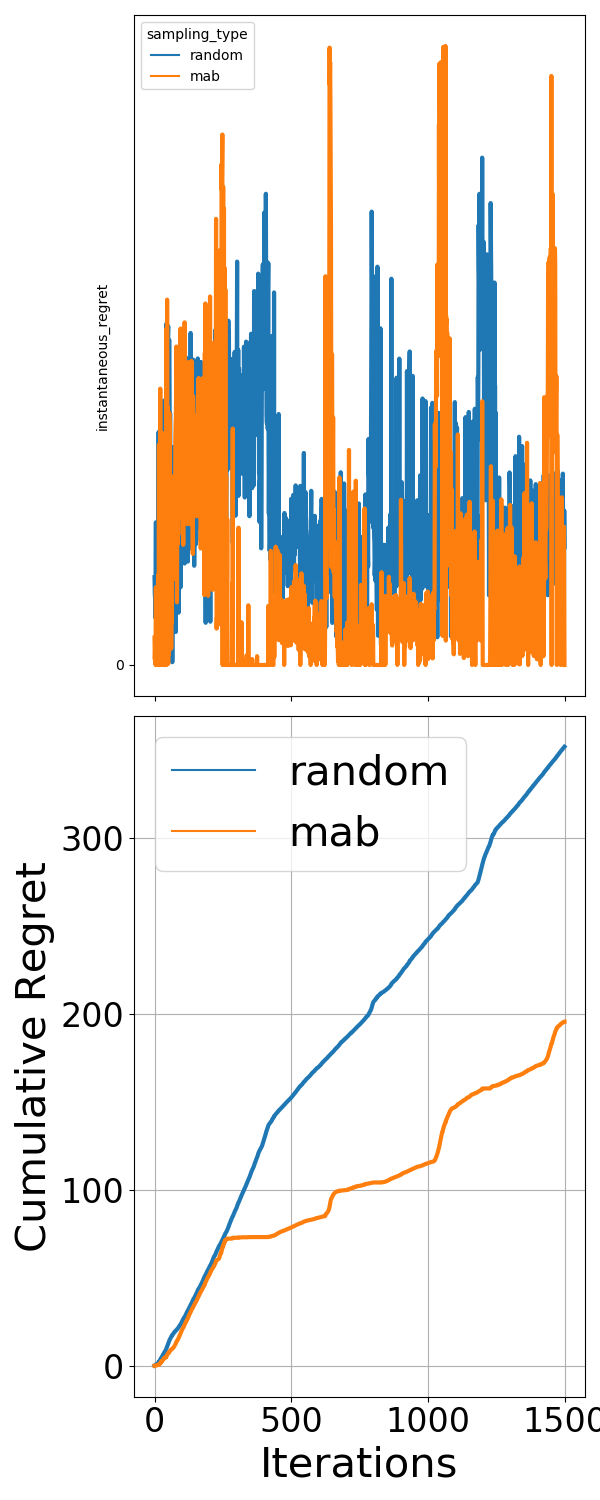}\label{fig: victor-results: regret}}
	\caption{Experimental results of the manipulation scenario.}
	\label{fig: victor-results}
 \vspace{-0.2cm}
\end{figure}

\begin{figure*}[tpb]
	\centering
    %\subfloat[][MAB-RRT]
    {\includegraphics[trim = 10cm 1cm 8cm 1cm, clip, angle=0, width=0.17\textwidth]{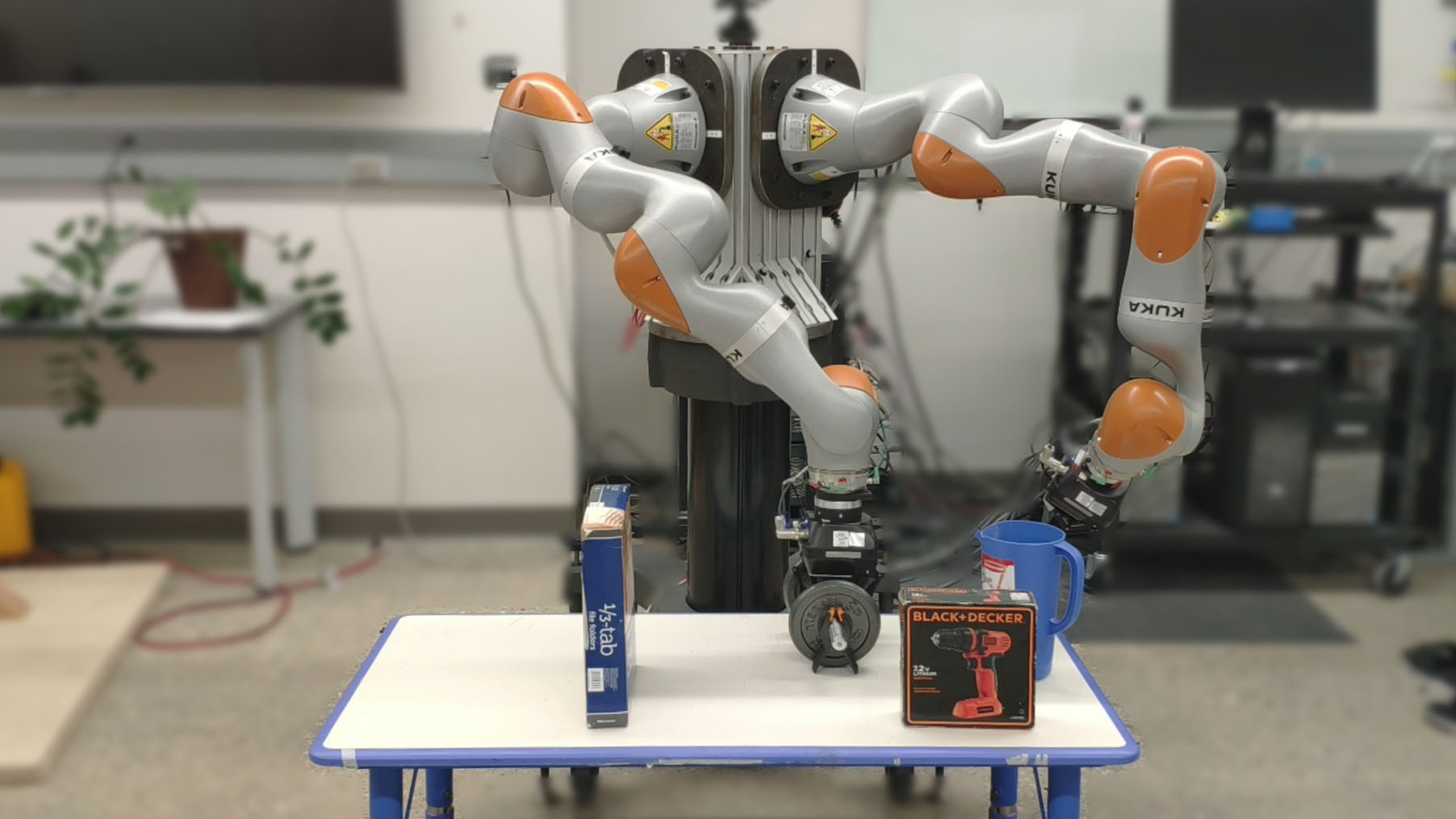}\,
    \includegraphics[trim = 10cm 1cm 8cm 1cm, clip, angle=0, width=0.17\textwidth]{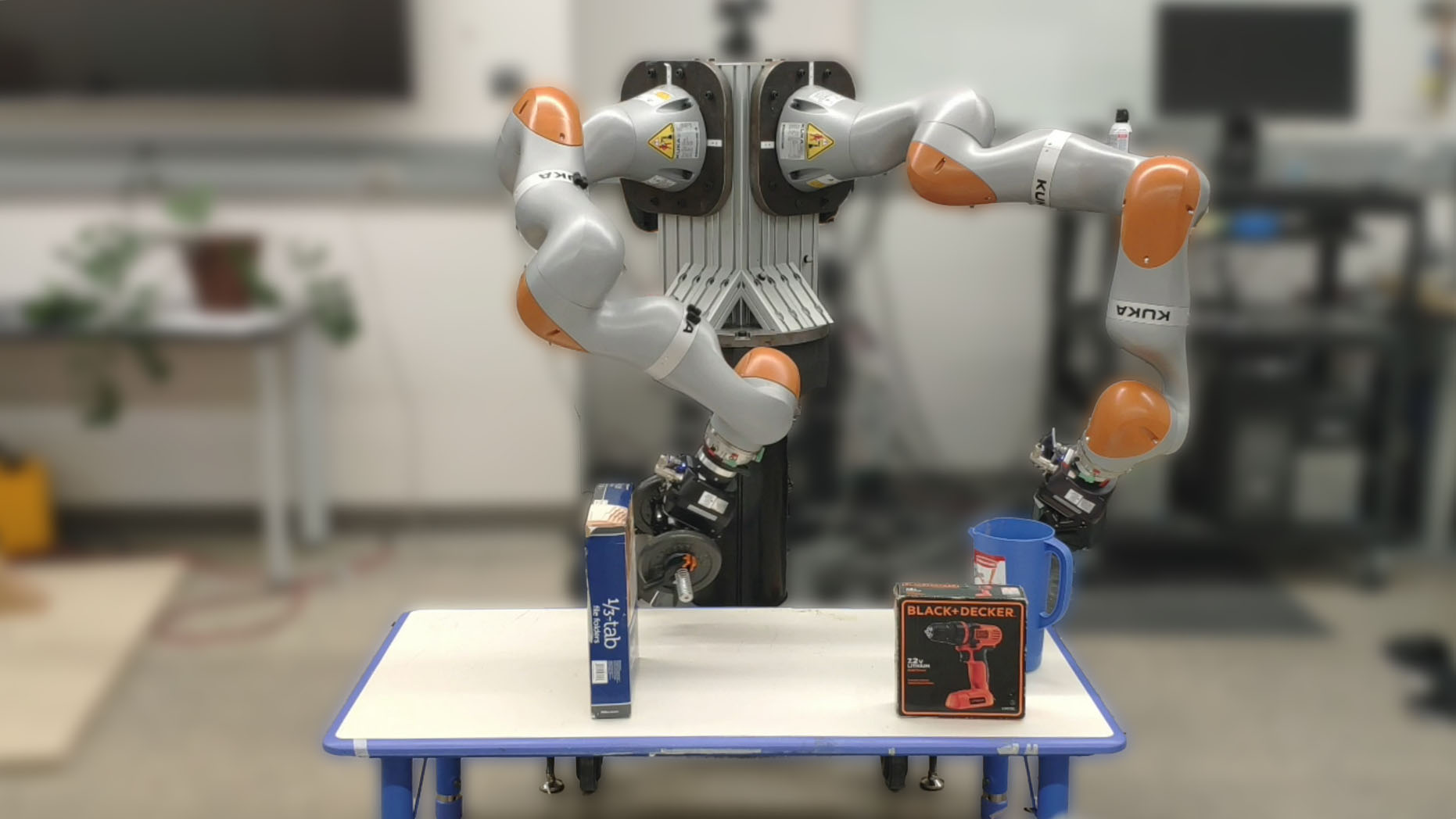}\,
    \includegraphics[trim = 10cm 1cm 8cm 1cm, clip, angle=0, width=0.17\textwidth]{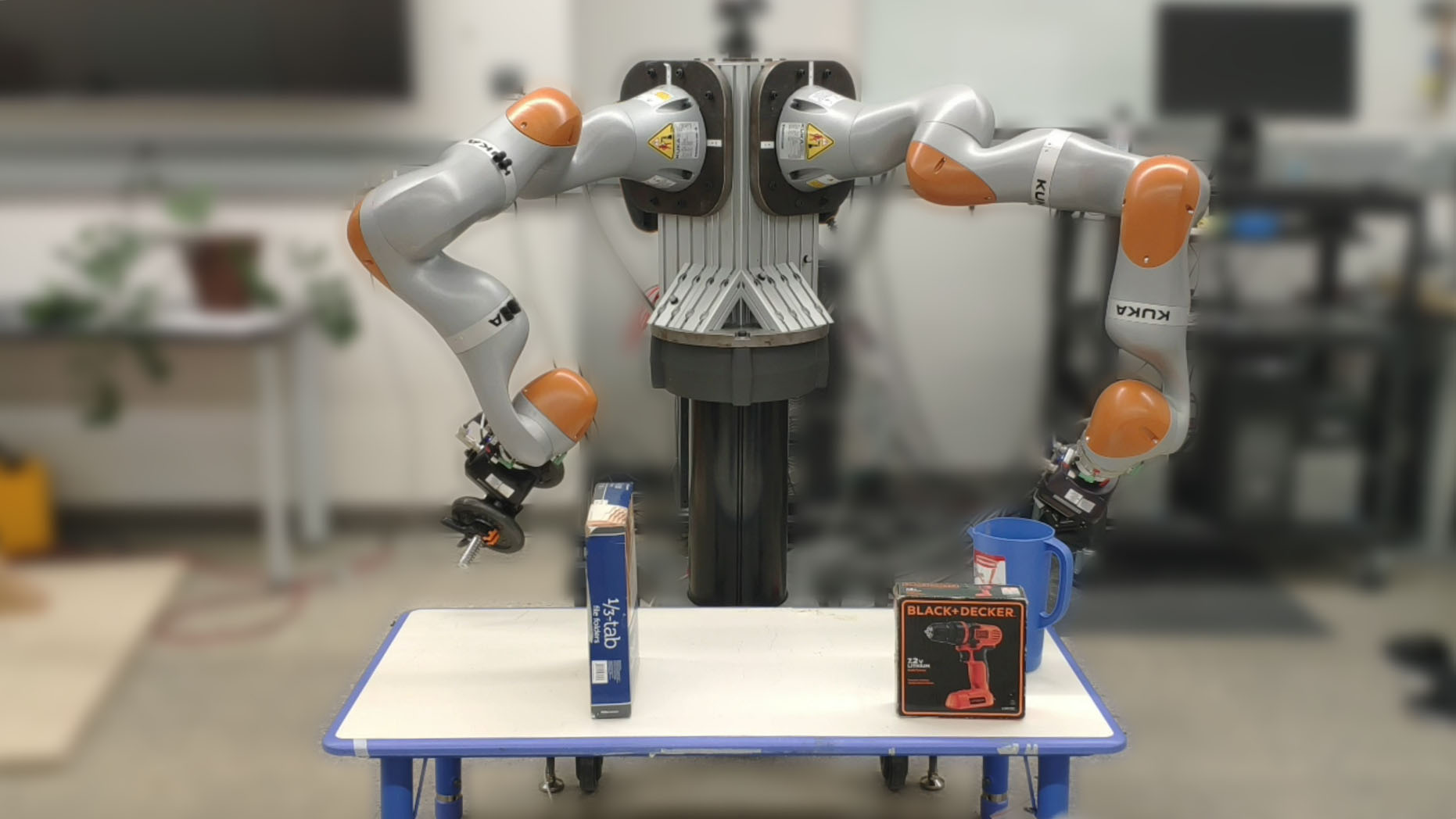}\,
    \includegraphics[trim = 10cm 1cm 8cm 1cm, clip, angle=0, width=0.17\textwidth]{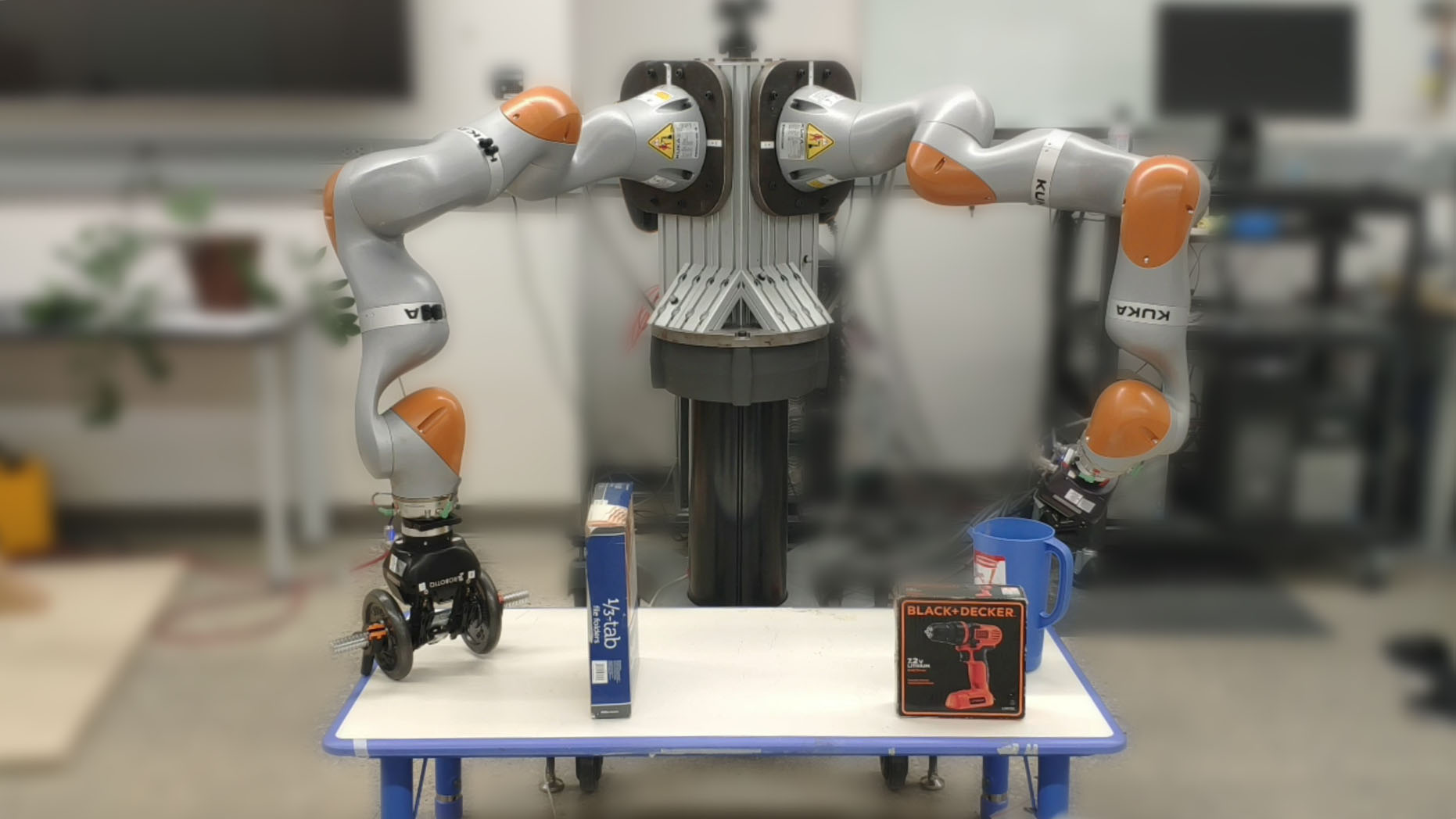}}\\
    \vspace{0.3cm}
    %\subfloat[][AO-RRT]
    {\includegraphics[trim = 10cm 1cm 8cm 1cm, clip, angle=0, width=0.17\textwidth]{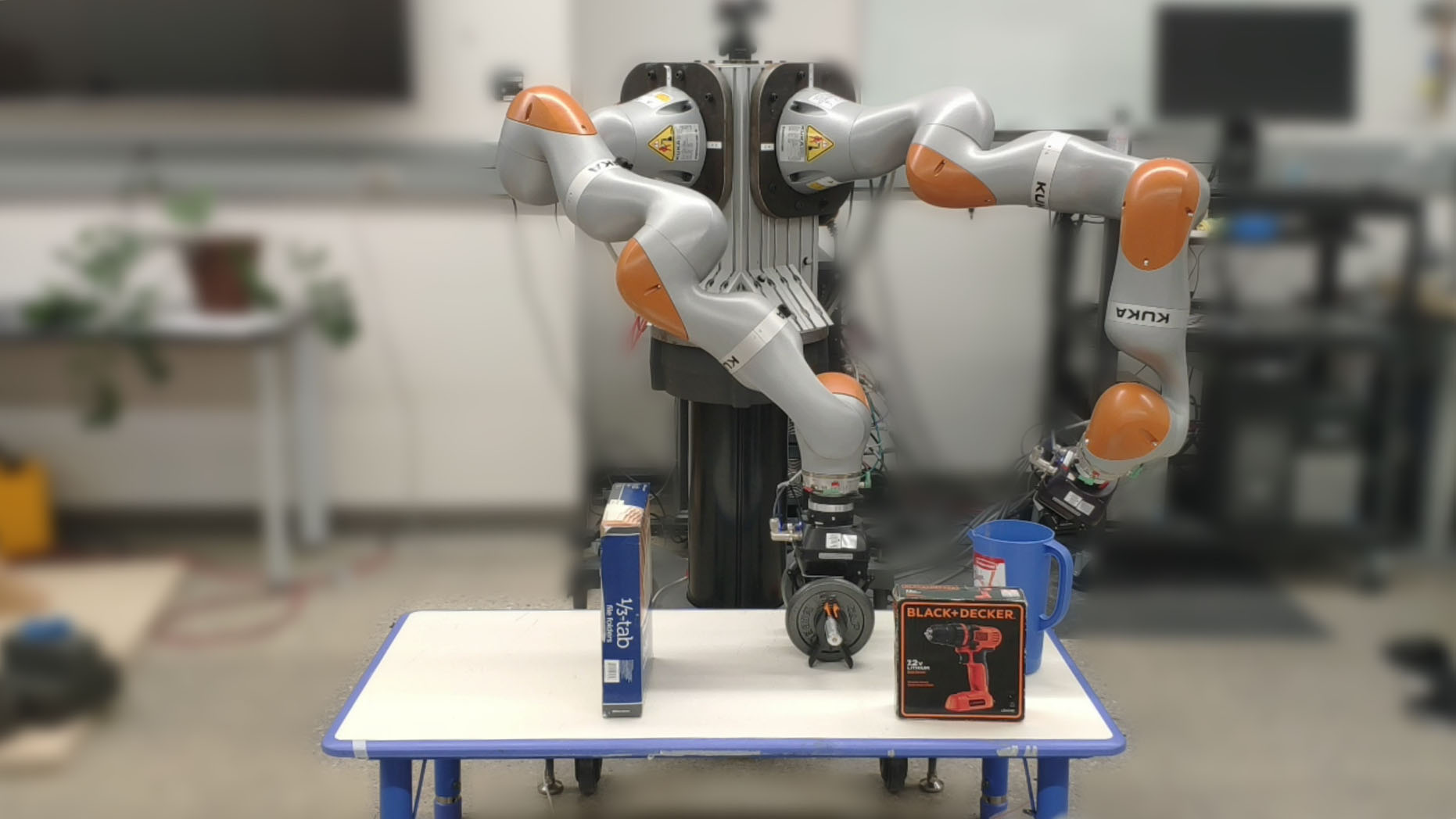}\,
    \includegraphics[trim = 10cm 1cm 8cm 1cm, clip, angle=0, width=0.17\textwidth]{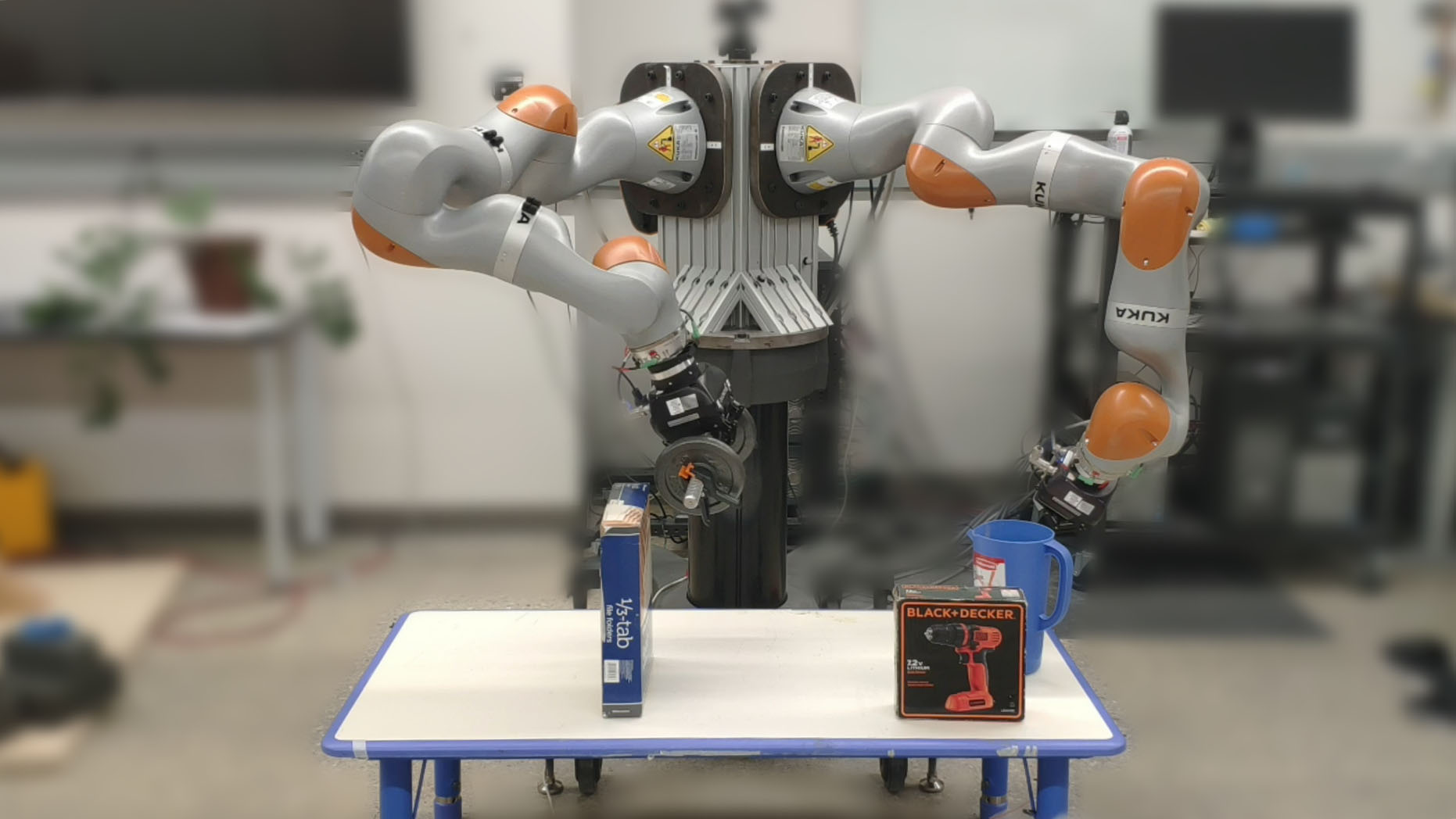}\,
    \includegraphics[trim = 10cm 1cm 8cm 1cm, clip, angle=0, width=0.17\textwidth]{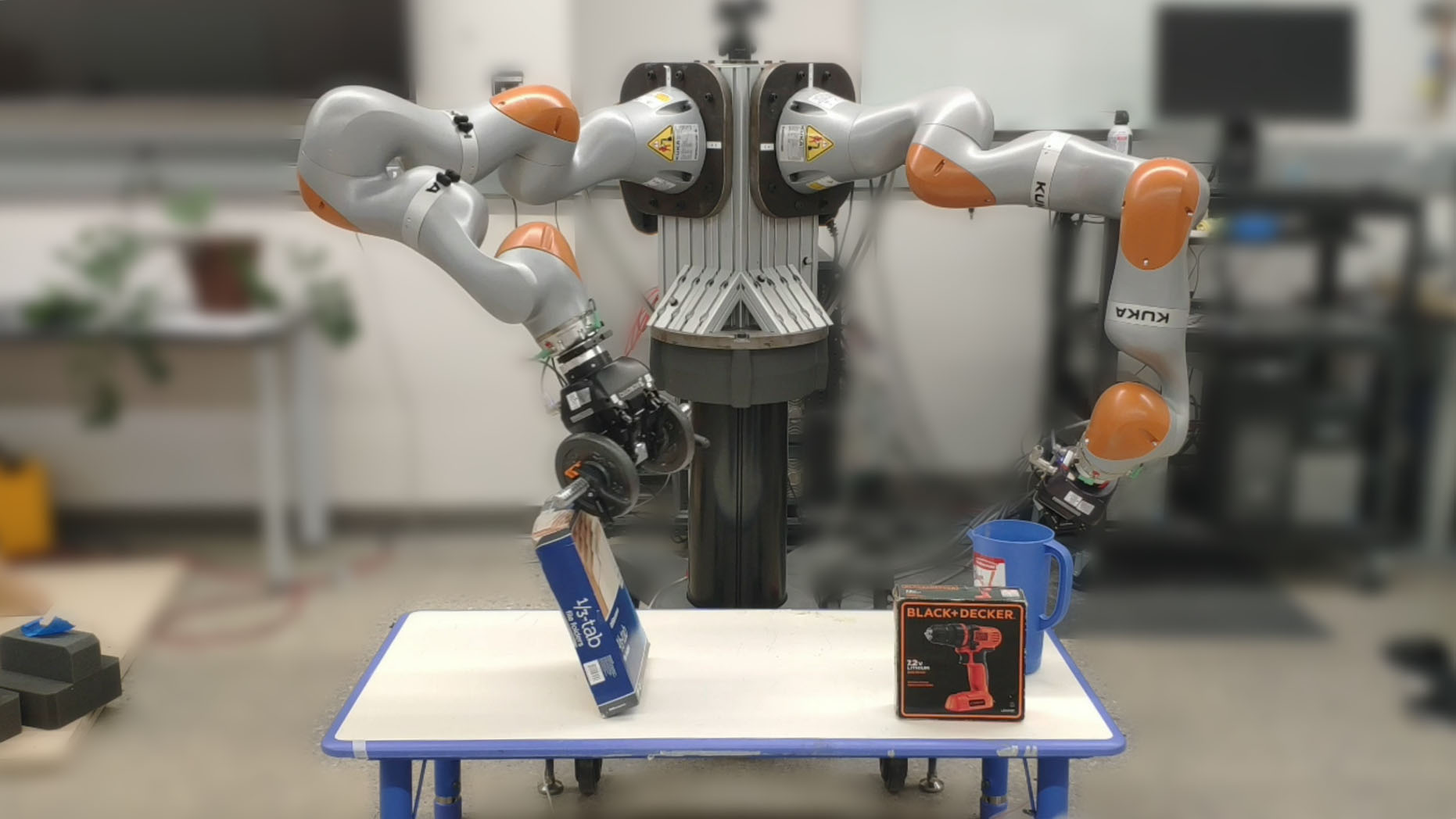}\,
    \includegraphics[trim = 10cm 1cm 8cm 1cm, clip, angle=0, width=0.17\textwidth]{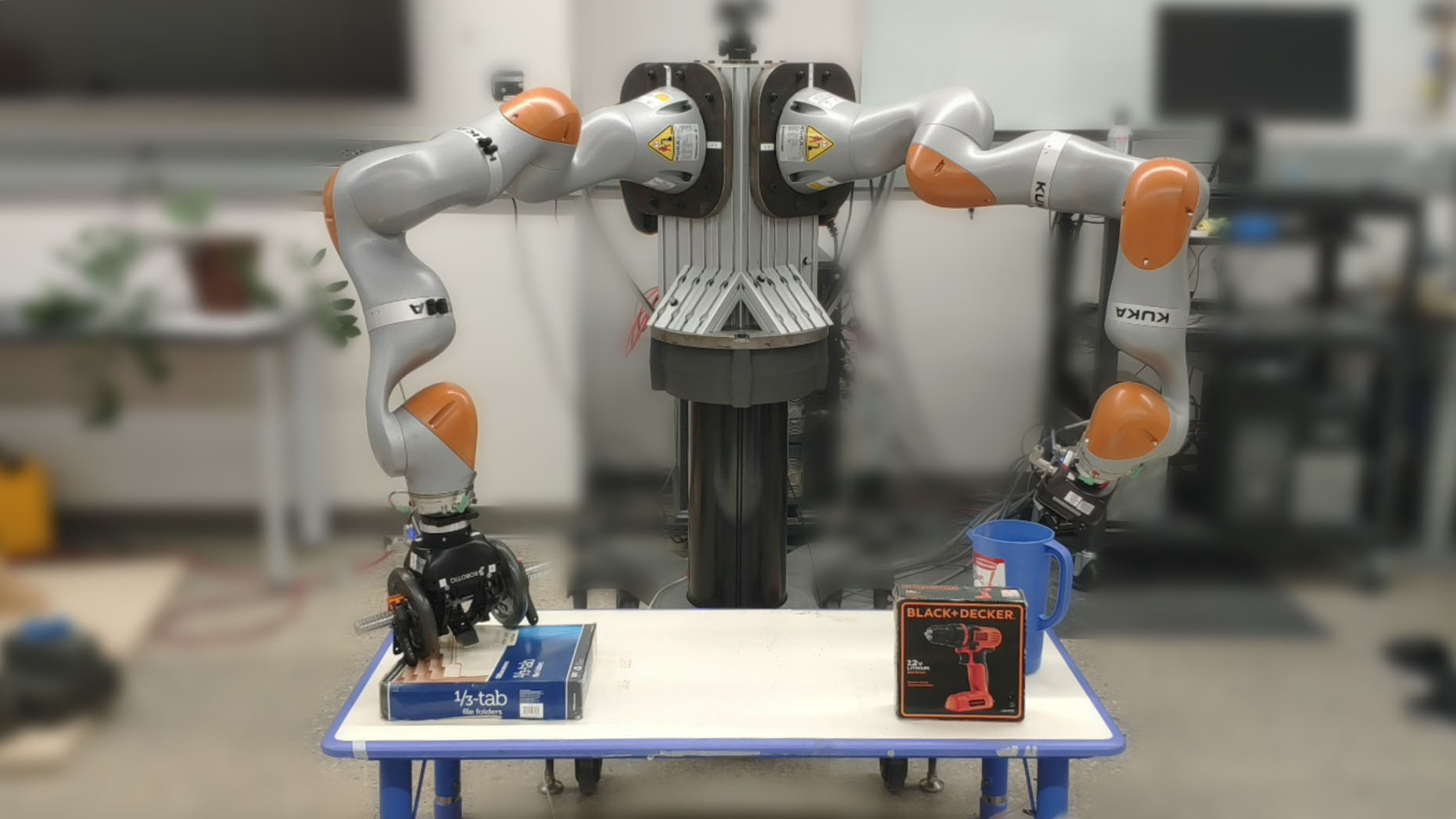}}
	\caption{Examples of executions of two trajectories planned with MAB-RRT (top) and AO-RRT (bottom).}
	\label{fig: victor-video-frames}
 \vspace{-0.5cm}
\end{figure*}

\bibliographystyle{IEEEtran}
\bibliography{bib,bib_new}

\end{document}